\documentclass[11pt]{article}
\usepackage{acl2016}
\usepackage{times}
\usepackage{latexsym}
\usepackage{graphicx}	
\usepackage{comment}
\usepackage{amsmath}
\usepackage{amssymb}
\usepackage{booktabs}
\usepackage{subcaption}
\usepackage{wrapfig}
\usepackage{url}
\usepackage[toc,page]{appendix}
\usepackage{enumitem}
\usepackage[toc,page]{appendix}
\setlist[itemize]{leftmargin=*}

\DeclareRobustCommand\onedot{\futurelet\@let@token\@onedot}
\def\@onedot{\ifx\@let@token.\else.\null\fi\xspace}

\graphicspath{{../iclr2016_sherlock/retrieval_qual3_seen_crop/}{../iclr2016_sherlock/retrieval_qual3_unseen_Crop/}{../iclr2016_sherlock/KEx_qual2_crop/}{../iclr2016_sherlock//KEx_qual2_unseen_crop/}{../iclr2016_sherlock/ret_qual3_gt5_seen_crop/}{../cvpr2016AuthorKit/figs/}
{../iclr2016_sherlock/}
{../iclr2016_sherlock_v2/}
 }

\aclfinalcopy 
\def\aclpaperid{264} 

\title{Automatic Annotation of Structured Facts in Images}

\author{ Mohamed Elhoseiny$^{1,2}$,  Scott Cohen$^{1}$,  Walter Chang$^{1}$,  Brian Price$^{1}$, Ahmed  Elgammal$^{2}$ \normalsize \\
$^{1}$Adobe Research $\,\,\,\,\,\,\,\,\,\,\,\,\,\,\,\,\,\,\,\,\,\,\,\,\,\,\,\,\,\,\,$ $^{2}$Department of Computer Science, Rutgers University}

\date{}

\begin{document}

\maketitle

\begin{abstract}
Motivated by the application of fact-level image understanding, we present an automatic method for data collection of structured visual facts from images with captions. Example structured facts include attributed objects (e.g., $<$flower, red$>$), actions (e.g., $<$baby, smile$>$), interactions (e.g., $<$man, walking, dog$>$), and positional information (e.g., $<$vase, on, table$>$). The collected annotations are in the form of fact-image pairs (e.g.,$<$man, walking, dog$>$ and an image region containing this fact). With a language approach, the proposed method is able to collect hundreds of  thousands of visual fact annotations  with accuracy of 83\% according to human judgment. 
Our method automatically collected more than 380,000 visual fact annotations and more than 110,000 unique visual facts from  images with captions and localized them in images in less than one day of processing time on standard CPU platforms. 
\end{abstract}

\section{Introduction}
People generally acquire visual knowledge by exposure
to both visual facts and to semantic or language-based
representations of these facts, e.g., by seeing
an image of ``a person petting dog'' and observing
this visual fact associated with its language
representation .
In this work,
we focus on methods for collecting structured facts that we
define as structures that provide attributes about an object, and/or
the actions and interactions this object may have with other objects. We introduce the idea of automatically collecting annotations
for second order visual facts and third order visual facts
where second order facts $<$S,P$>$ are attributed objects
(e.g., $<$S: car, P: red$>$) and single-frame actions (e.g.,
$<$S: person, P: jumping$>$), and third order facts specify
interactions (i.e., $<$boy, petting, dog$>$). 
This structure is helpful for designing
machine learning algorithms that learn deeper image semantics
from caption data and allow us to model the relationships between facts.
In order to enable such a setting, we
need to collect these structured fact annotations
in the form of (language view, visual view) pairs
(e.g., $<$baby, sitting on, chair$>$ as the language
view and an image with this fact as a visual view)
to train models.


\begin{figure*}[t!]
\vspace{-4mm}
  \centering
    \includegraphics[width=1.0\textwidth]{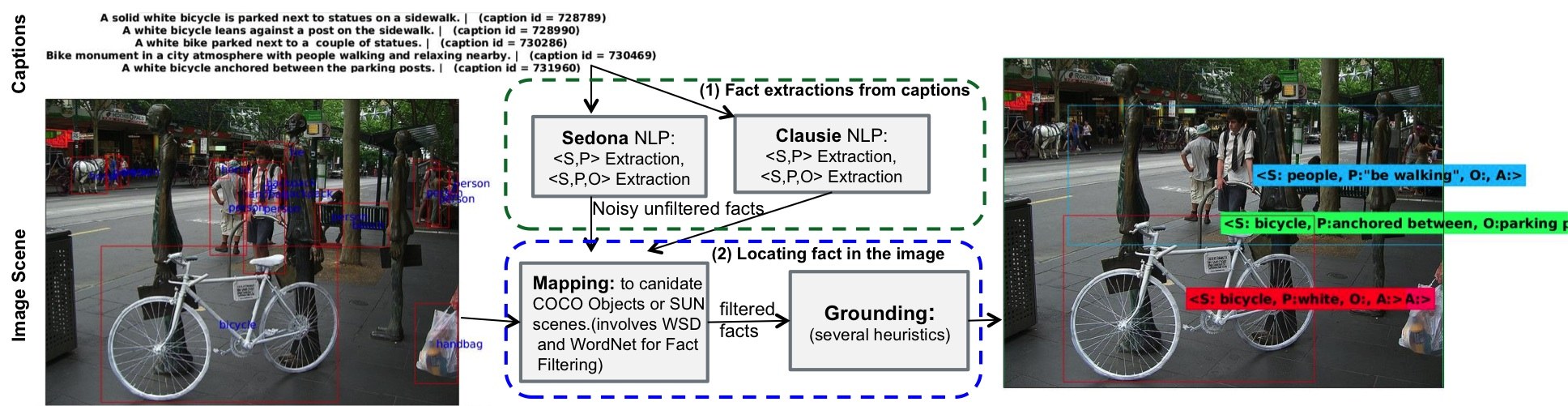}
      \caption{Structured Fact Automatic Annotation}
      \label{fig:afaf}
      \vspace{-2mm}
\end{figure*}


~\cite{chen2013neil}  showed that visual concepts, from a predefined ontology, can be learned by querying the web about these concepts using image-web search engines. More recently,~\cite{divvala2014learning} presented an approach to learn concepts related to a particular object by querying the web with Google-N-gram data that has the concept name.
There are three limitations to these approaches. 
(1) It is difficult to define the space of  visual knowledge and then search for it. It is further restricting to define it based on a predefined ontology such as~\cite{chen2013neil} or a particular object such as ~\cite{divvala2014learning}. 
(2) Using image search is not reliable to collect data for concepts with few images on the web. These methods assume that the top retrieved examples by image-web search are positive examples and that there are images available that are annotated with the searched concept.
(3) These concepts/facts are not structured and  hence annotations lacks information like ``jumping'' is the action part in $<$person, jumping $>$, or ``man' and ``horse'' are interacting  in  $<$person, riding, horse $>$. This structure is important for deeper understanding of visual data, which is one of the main motivations of this work. 

The problems in the prior work motivate us to propose a method to automatically annotate structured facts  by  processing image caption data since facts in image captions are highly likely to be located in the associated  images.  We show that  a large quantity of high quality structured visual facts could be extracted from caption datasets using natural language processing methods. Caption writing is free-form and an easier task for crowd-sourcing workers than labeling second- and third-order tasks, and such free-form descriptions are readily available in existing image caption datasets. We focused on collecting facts from the MS COCO image caption dataset~\cite{lin2014microsoft} and the newly collected Flickr30K entities~\cite{plummer2015flickr30k}. 
We automatically  collected more than 380,000 structured fact annotations in high quality from both the 120,000 MS COCO scenes and  30,000 Flickr30K scenes.


The main contribution of this paper is an accurate, automatic, and efficient method for extraction of structured fact visual annotations from image-caption datasets, as illustrated in Fig.~\ref{fig:afaf}. Our approach  (1) extracts facts from captions associated with images and then (2)  localizes the extracted facts in the image.
For fact extraction from captions,
We propose a new method called {\em SedonaNLP} for fact extraction 
to fill gaps in existing fact extraction from sentence methods like Clausie~\cite{del2013clausie}. 
SedonaNLP produces more facts than Clausie, especially $<$subject,attribute$>$ facts, and thus enables collecting more visual annotations than using Clausie alone.
The final set of automatic annotations are the set of successfully localized facts in the associated images.  
We show that these facts are extracted with more than 80\% accuracy according to human judgment.  


\section{Motivation}

Our goal by proposing this automatic method is to  generate language\&vision annotations at the fact-level to help study language\&vision for the sake of structured understanding of visual facts. Existing systems already work on relating captions directly to the whole image such as \cite{karpathy2014deep,kiros2014unifying,vinyals2015show,xu2015show,mao2015deep,antol2015vqa,malinowski2015ask,ren2015exploring}. This gives rise to a key question about our work: why it is useful to collect such a large quantity of structured facts compared to caption-level systems?   

We illustrate the difference between caption-level learning  fact-level learning that motivates this work  by the example in Fig~\ref{fig:afaf}. Caption-level  learning systems correlate captions like those on top of   Fig.~\ref{fig:afaf}(top-left) to the whole image that includes all objects. Structured Fact-level learning systems  are instead fed with localized annotations for each fact extracted  form the image caption; see in Fig.~\ref{fig:afaf}(right), Fig.~\ref{safa_pos_Ex}, and ~\ref{safa_neg_Ex} in Sec.~\ref{sec_Expr}. 
Fact level annotations are less confusing training data than sentences 
because they provide more precise information for 
both the language and the visual views. \textbf{(1)}  From the language view, the annotations we generate is precise to list a particular fact (e.g., $<$bicycle,parked between, parking posts$>$). \textbf{(2)} From the visual view, it provide the bounding box of this fact; see Fig~\ref{fig:afaf}. \textbf{(3)} A third unique part about our annotations is the structure: e.g., $<$bicycle,parked between, parking posts$>$ instead of ``a bicycle parked between parking posts''. 

Our collected data has been used to develop methods that  learn hundreds of thousands of image facts, as we introduced and studied in~\cite{sherlock_eccv16}.  The results shows that  fact-level learning is superior compared to caption-level learning like~\cite{kiros2014unifying}, as shown in Table~4 in~\cite{sherlock_eccv16} (16.39\% accuracy versus  3.48\% for ~\cite{kiros2014unifying}).  It further shows the value of the associated structure in the  (16.39\% accuracy  versus 8.1\%) in Table~4\cite{sherlock_eccv16}).  Similar results also shown on a smaller scale in Table~3 in~\cite{sherlock_eccv16}.  

\section{Approach Overview}
\label{ss:dc_2}
    \vspace{-1.5mm}

We propose a two step automatic annotation of structured facts: 
(i) Extraction of structured fact  from captions, and (ii) Localization of these facts in images. First, the captions associated with the given image are analyzed to extract sets of clauses that are considered as candidate $<$S,P$>$, and $<$S,P,O$>$ facts. 

Captions can provide a tremendous amount of information to image understanding systems. However, developing NLP systems to accurately and completely extract structured knowledge from free-form text 
is an open problem. 
We extract structured facts using two methods: Clausie~\cite{del2013clausie} and  Sedona( detailed later in Sec~\ref{sec_fact_extract}); also see Fig~\ref{fig:afaf}. We found Clausie~\cite{del2013clausie} missed many visual facts in the captions which motivated us to develop Sedona to fill this gap as detailed in Sec.~\ref{sec_fact_extract}. 

Second, we  localize these facts within the image (see Fig.~\ref{fig:afaf}). The successfully located facts in the images are saved as fact-image annotations that could be used to train visual perception models to learn attributed objects, actions, and interactions. We managed to collect 380.409 high-quality second- and third-order fact annotations (146,515 from Flickr30K Entities, 157,122 from the MS COCO training set, and 76,772 from the MS COCO validation set). 
We present  statistics of the automatically collected facts in the Experiments section.  
Note that the process of localizing facts in an image is constrained by information in the dataset.

 For MS COCO, the dataset contains object annotations for about 80 different objects as provided by the training and validation sets. Although this provides abstract information about objects in each image (e.g., ``person''), it is usually mentioned in different ways in the caption. For the  ``person'' object, ``man'', ``girl'', ``kid'', or ``child'' could instead appear in the caption.  In order to locate second- and third-order facts in images, we started by defining visual entities. For the MS COCO dataset~\cite{lin2014microsoft}, we define a visual entity as any noun that is either (1) one of the MS COCO dataset objects,  (2) a noun in the WordNet ontology ~\cite{miller1995wordnet,leacock1998combining} that is an immediate or indirect hyponym of one of the MS COCO objects (since WordNet is searchable by a sense and not a word, we perform word sense disambiguation on the sentences using a state-of-the-art method~\cite{zhong2010makes}), or (3) one of scenes the SUN dataset~\cite{xiao2010sun} (e.g., a ``restaurant'').  We expect visual entities to appear either in the S or the O part (if exists) of a candidate fact. This allows us to then localize facts for images in the MS COCO dataset. Given a candidate third-order fact, we first try to assign each S and O to one of the visual entities. If S and O elements are not visual entities, then the fact is ignored.  Otherwise,  
the facts are processed by several heuristics, detailed in Sec~\ref{sec_locate_facts}. 
For instance, our method takes into account that grounding the plural "men" in the fact $<$S:men, P: chasing, O: soccer ball $>$ may require the union of multiple "man" bounding boxes.

In the Flickr30K Entities dataset~\cite{plummer2015flickr30k}, the bounding box annotations are presented as phrase labels for sentences (for each phrase in a caption that refers to an entity in the scene).  
A visual entity is considered to be a phrase with a bounding box annotation or one of the SUN scenes. Several heuristics were developed and applied to collect these fact annotations, e.g. grounding a fact about a scene to the entire image; detailed in Sec~\ref{sec_locate_facts}.

\section{Fact Extraction from Captions}
\label{sec_fact_extract}
\begin{figure}[t!]
\begin{center}
\includegraphics[width=0.5\textwidth]{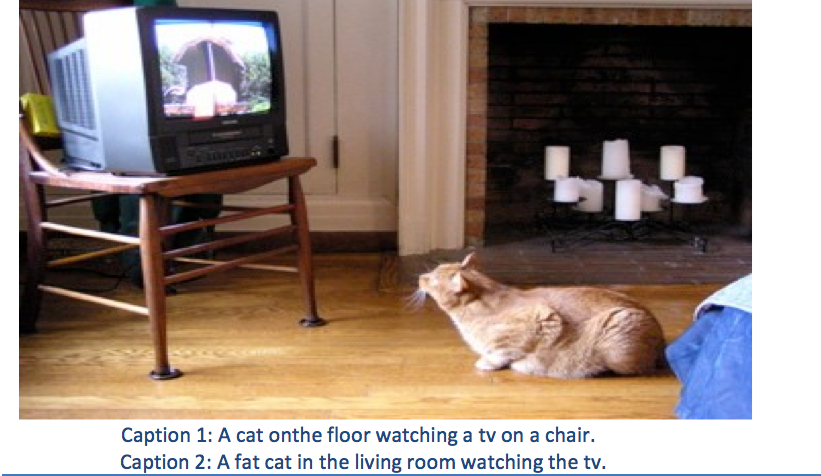}
\includegraphics[width=0.5\textwidth]{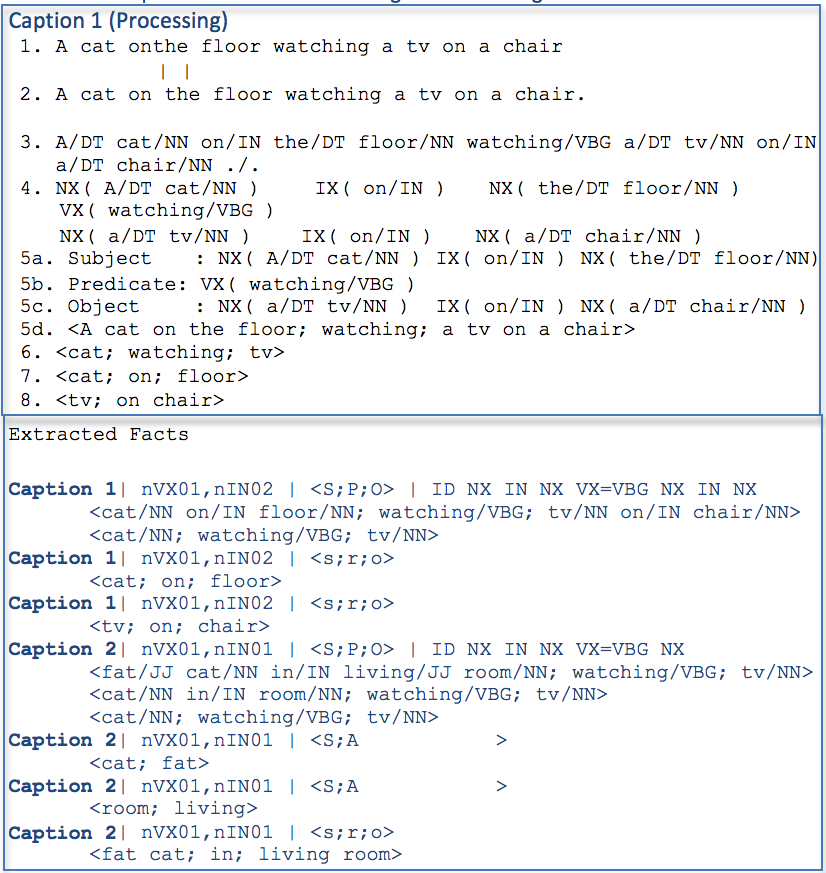}
\end{center}
\vspace*{-5mm}
\caption{Examples of caption processing and $<$S,P,O$>$ and $<$S,P$>$ structured fact extractions.}
\label{fig_sedona_ex}
\end{figure}

We extract facts from captions using Clausie~\cite{del2013clausie} and our proposed SedonaNLP system. 
In contrast to Clausie, we address several challenging linguistic issues by evolving our NLP pipeline to: 1) correct many common spelling and punctuation mistakes, 2) resolve word sense ambiguity within clauses,
 and 3) learn a common spatial preposition lexicon (e.g., ``next\_to'', ``on\_top\_of'', ``in\_front\_of'') that consists of over 110 such terms, as well as a lexicon of over two dozen collection phrase adjectives (e.g., 
"group\_of", 
"bunch\_of", 
"crowd\_of", 
"herd\_of"). 
For our purpose, these strategies allowed us to extract more interesting structured facts that Clausie fails at which include (1) more discrimination between single versus plural terms, (2) extracting positional facts (e.g., next\_to). Additionally, SedonaNLP produces attribute facts that we denote as $<$S, A$>$; see Fig~\ref{fig_sedona_ex}. 
Similar to some existing systems OpenNLP~\cite{baldridge2014opennlp} and ClearNLP~\cite{clearNLP14}, the SedonaNLP platform also performs many common NLP tasks: e.g., sentence segmentation, tokenization, part-of-speech tagging, named entity extraction, chunking, dependency and constituency-based parsing, and coreference resolution.  SedonaNLP itself employs both open-source components such as NLTK and WordNet, as well as internally-developed annotation algorithms for POS and clause tagging.  These tasks are used to create more advanced functions such as structured fact annotation of images via semantic triple extraction.  In our work, we found SedonaNLP and Clausie to be complementary for producing a set of candidate facts for possible localization in the image that resulted in successful annotations. 


\begin{figure*}[htp]
\begin{center}
\includegraphics[width=0.95\textwidth]{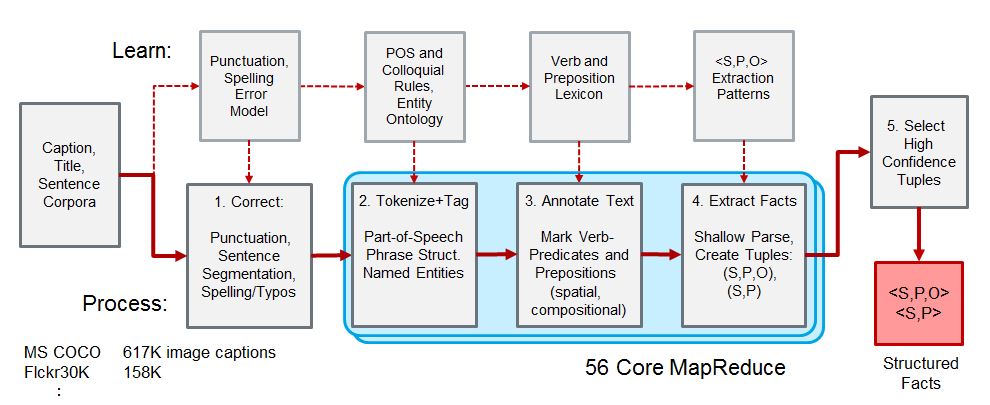}
\end{center}
\vspace*{-5mm}
\caption{SedonaNLP Pipeline for Structured Fact Extraction from Captions}
\label{fig-SedonaNLP}
\end{figure*}

\begin{figure}[t!]
\begin{center}
\includegraphics[width=0.45\textwidth]{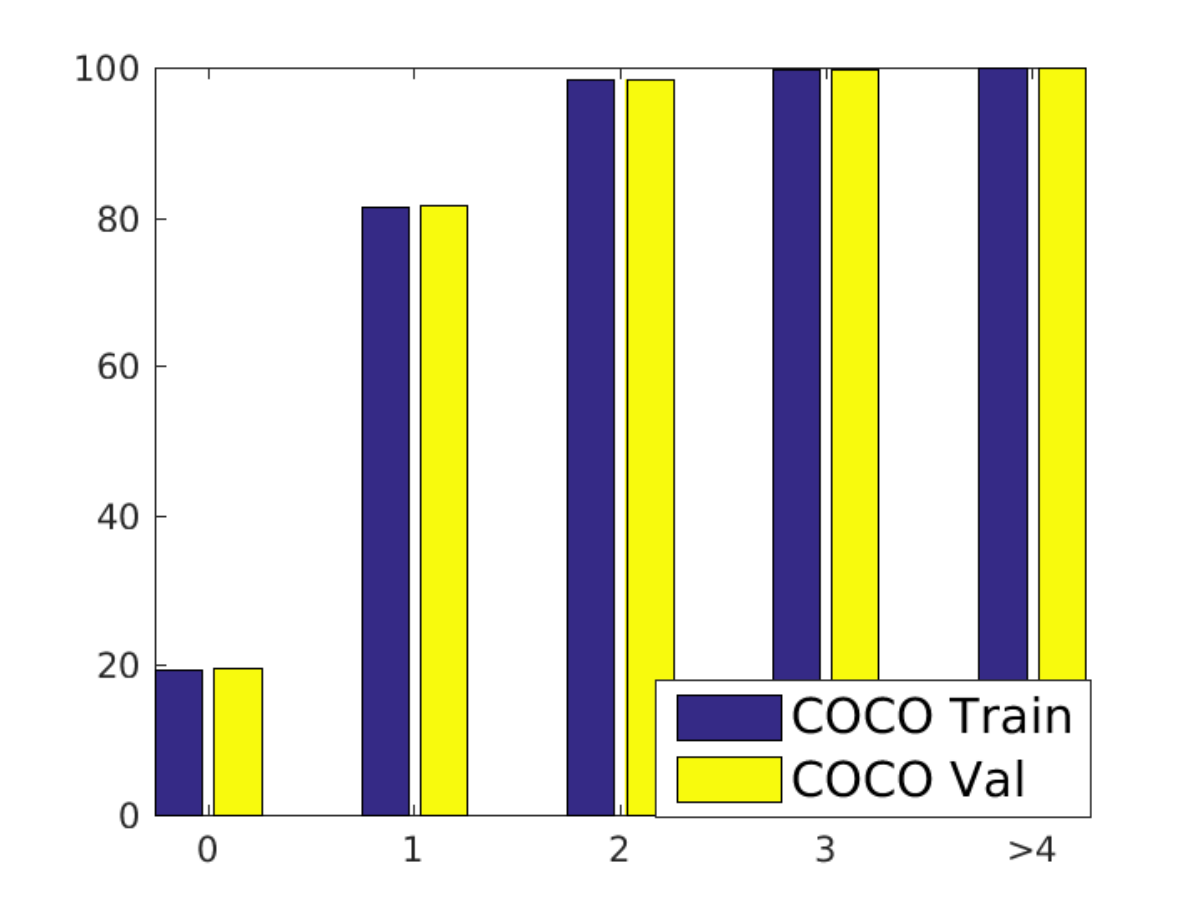}
\end{center}
\vspace*{-5mm}
\caption{Accumulative Percentage of SP and SPO facts in COCO 2014 captions as number of verbs increases}
\label{fig-COCOverbs}
\end{figure}

\begin{figure*}[hbtp!]
\begin{center}
\includegraphics[width=0.95\textwidth]{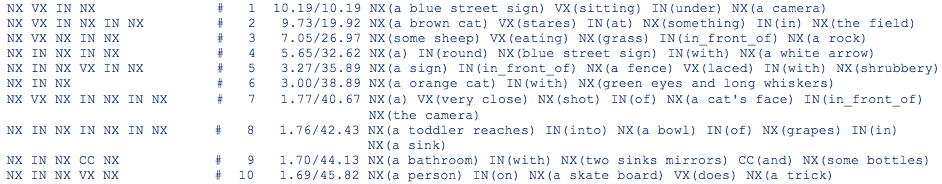}
\end{center}
\vspace*{-5mm}
\caption{Examples of the top observed Noun (NX), Verb (VX), and Preposition (IN) Syntactic patterns.}
\label{fig_top10}
\end{figure*}

Varying degrees of success have been achieved in extracting and representing structured triples from sentences using $<$subject, predicate, object$>$ triples.  For instance,~\cite{rusu2007triplet} describe a basic set of methods based on traversing the parse graphs generated by various commonly available parsers. Larger scale text mining methods for learning structured facts for question answering have been developed in the IBM Watson PRISMATIC framework~\cite{fan2010prismatic}.   
While parsers such as CoreNLP~\cite{manning2014stanford} are available to generate comprehensive dependency graphs, these have historically required significant processing time for each sentence or have traded accuracy for performance.
In contrast, SedonaNLP currently employs a shallow dependency parsing method that runs in some cases  8-9X faster than earlier cited methods running on identical hardware.  We choose a shallow approach with high, medium, and low confidence cutoffs after observing that roughly 80\% of all captions consisted of 0 or 1 Verb expressions (VX); see Fig.~\ref{fig-COCOverbs} for MSCOCO dataset~\cite{lin2014microsoft}.  The top 500 image caption syntactic patterns we observed can be found on our supplemental materials.  These syntactic patterns are used to learn rules for automatic extraction for not only  $<$S,P,O$>$, but also $<$S,P$>$, and  $<$S,A$>$, where $<$S,P$>$, are subject-action facts and $<$S,A$>$ are subject-attribute facts. Pattern examples and statistics for MS COCO are shown in Fig.~\ref{fig_top10}.

In SedonaNLP, structured fact extraction was accomplished by learning a subset of abstract syntactic patterns consisting of basic noun, verb, and preposition expressions by analyzing  1.6M caption examples provided by the MS COCO, Flickr30K, and Stony Brook University Im2Text caption datasets.  Our approach mirrors existing known art with the addition of internally-developed POS and clause tagging accuracy improvements through the use of heuristics listed below to reduce higher occurrence errors due to systematic parsing errors: (i) Mapping past participles to adjectives (e.g., stained glass), (ii) De-nesting existential facts (e.g., this is a picture of a cat watching a tv.), (iii) Identifying auxiliary verbs (e.g.,  do verb forms).

In Fig.~\ref{fig_sedona_ex}, we show an example of extracted $<$S,P,O$>$ structured facts useful for image annotation for a small sample of MS COCO captions. Our initial experiments empirically confirmed the findings of IBM Watson PRISMATIC researchers who indicated big complex parse trees tend to have more wrong parses. By limiting a frame to be only a small subset of a complex parse tree, we reduce the chance of error parse in each frame~\cite{fan2010prismatic}.  In practice, we observed many correctly extracted structured facts  for the more complex sentences (i.e., sentences with multiple VX verb expressions and multiple spatial prepositional expressions) -- these facts contained useful information that could have been used in our joint learning model but were conservatively filtered to help ensure the overall accuracy of the facts being presented to our system. As improvements are made to semantic triple extraction and confidence evaluation systems, we see potential in several areas to exploit more structured facts and to filter less information.  Our full $<$S,P,O$>$ triple and related tuple extractions for MS COCO and Flickr30K datasets are available in the supplemental material.

\section{Locating facts in the Image}
\label{sec_locate_facts}
In this section, we present details about the second step of our automatic annotation process introduced in Sec.~\ref{ss:dc_2}. 
After the candidate facts are extracted from the sentences, 
we end up with a set $\mathbf{F}_s = \{ \mathbf{f}_l^i \}, i = 1:N_s$ for statement $s$, where $N_s$ is the number of extracted  candidate fact  $\mathbf{f}_l^i, \forall i$ from the statement $s$ using either Clausie~\cite{del2013clausie} or Sedona-3.0.  The localization step is further divided into two steps. The mapping step maps nouns in the facts to candidate boxes in the image.  The grounding step processes each fact associated with the candidate boxes and outputs a final bounding box if localization is successful. The two steps are detailed in the following subsections.

\subsection{ Mapping}

The mapping step starts with a pre-processing step that filters out a non-useful subset of $\mathbf{F}_s$ and produces a more useful set  $\mathbf{F}_s^*$ that we try to locate/ground in the image. We perform this step by performing word sense disambiguation using the state-of-the-art method  ~\cite{zhong2010makes}. The word sense disambiguation method provides each word in the statement with a word sense in the wordNet ontology~\cite{leacock1998combining}. It also assigns for each word a part of speech tag. Hence, for each extracted candidate fact in $\mathbf{F}_s$ we can verify  if it follows the expected part of speech according to~\cite{zhong2010makes}.  For instance, all S should be nouns, all P should be either verbs or adjectives, and O should be nouns.  This results in a filtered set of facts $\mathbf{F}_s^*$. Then, each S is associated with a set of candidate boxes in the  image for second- and third-order facts and each O associated with a set or candidate boxes in the image for third-order facts only.  Since entities in MSCOCO dataset and Flickr30K are annotated differently, we present how the candidate boxes are determined in each of these datasets.

\textbf{MS COCO Mapping: } Mapping to candidate boxes for MS COCO reduces to assigning the S for second-order and third-order facts, and S and O for third-order facts.  Either S or O is assigned to one of the MSCOCO objects or SUN scenes classes. Given the word sense of the given part (S or O), we check if the given sense is a descendant of MSCOCO objects senses in the wordNet ontology. If it is, the given part (S or O) is associated with the set of candidate bounding boxes that belongs to the given object (e.g., all boxes that contain the  ``person'' MSCOCO object is under the  ``person'' wordnet node like ``man'', 'girl', etc). If the given part (S or O) is not an MSCOCO object or one of its descendants under wordNet, we further check if the given part is one of the SUN dataset scenes. If this condition holds, the given part is associated with a bounding box of the whole image.

\textbf{Flickr30K Mapping: } 
In contrast to MSCOCO dataset, the bounding box annotation comes for each entity in each statement in Flickr30K dataset. Hence, we compute the candidate bounding box annotations for each candidate fact by searching the entities in the same statement from which the clause is extracted. Candidate boxes are those that have the same name. Similarly, this process  assigns  S  for second-order facts and assigns S and O for second- and third-order facts.

Having finished the mapping process, whether for MSCOCO or Flickr30K, each candidate fact $\mathbf{f}_l^i \in \mathbf{F}_s^*$, is associated with candidate boxes depending on its type as follows.

\textbf{$<$S,P$>$ : } Each $\mathbf{f}_l^i \in  \mathbf{F}_s^*$ of second-order type is associated with one set of bounding boxes $\mathbf{b}_S^i$, which are the candidate boxes for the S part. $\mathbf{b}_O^i$ could be assumed to be always an empty set for second-order facts. 

\textbf{$<$S,P,O$>$ : } Each $\mathbf{f}_l^i \in  \mathbf{F}_s^*$ of third-order type is associated with two sets of bounding boxes $\mathbf{b}_S^i$ and $\mathbf{b}_S^i$ as  candidate boxes for the S and P parts, respectively. 

\subsection{Grounding}

The grounding process is the process of associating each $\mathbf{f}_l^i \in \mathbf{F}_s^*$ with an image $\mathbf{f}_v$ by assigning $\mathbf{f}_l$ to a bounding box in the given MS COCO image scene given the  $\mathbf{b}_S^i$ and $\mathbf{b}_O^i$ candidate boxes.  The grounding process is relatively different for the two dataset due to the difference of the entity annotations. 

\textbf{Grounding: MS COCO dataset (Training and Validation sets)}
\begin{table*}[htbp!]
  \centering
\caption{Human Subject Evaluation by MTurk workers \% }
\label{tbl:afaf_eval_mturk}
\scalebox{0.8}
{
\begin{tabular}{|c|cc|cc|ccccccc|}
    \hline
\textbf{  Dataset (responses) }       & \multicolumn{2}{c|}{\textbf{Q1}} & \multicolumn{2}{c|}{\textbf{Q2}} & \multicolumn{7}{c|}{\textbf{Q3}} \\
\hline
          & \textbf{yes} & \textbf{no} & \textbf{Yes} & \textbf{No} & \textbf{a} & \textbf{b} & \textbf{c} & \textbf{d} & \textbf{e} & \textbf{f} & \textbf{g} \\ \hline
    \textbf{MSCOCO train 2014 (4198)} & 89.06 & 10.94 & 87.86 & 12.14 & 64.58 & 12.64 & 3.51  & 5.10  & 0.86  & 1.57  & 11.73 \\
    \textbf{MSCOCO val 2014 (3296)} & 91.73 & 8.27  & 91.01 & 8.99  & 66.11 & 14.81 & 3.64  & 4.92  & 1.00  & 0.70  & 8.83 \\
    \textbf{Flickr30K Entities2015 (3296)} & 88.94 & 11.06 & 88.19 & 11.81 & 70.12 & 11.31 & 3.09  & 2.79  & 0.82  & 0.39  & 11.46 \\
    \hline
    \textbf{Total} & 89.84 & 10.16 & 88.93 & 11.07 & 66.74 & 12.90 & 3.42  & 4.34  & 0.89  & 0.95  & 10.76 \\
\hline
    \end{tabular}
    }
\end{table*}

\begin{table*}[t!]
  \centering
\caption{Human Subject  Evaluation by Volunteers \% (This is another set of annotations different from those evaluated by MTurkers)}
\vspace{-4mm}
\label{tbl:afaf_eval_mturk}
\scalebox{0.8}
{
    \begin{tabular}{|c|cc|cc|ccccccc|}
   \hline
    \textbf{Volunteers}      & \multicolumn{2}{c|}{\textbf{Q1}} & \multicolumn{2}{c|}{\textbf{Q2}} & \multicolumn{7}{c|}{\textbf{Q3}} \\
  \hline
          & \textbf{yes} & \textbf{No} & \textbf{Yes} & \textbf{No} & \textbf{a} & \textbf{b} & \textbf{c} & \textbf{d} & \textbf{e} & \textbf{f} & \textbf{g} \\
    \textbf{MSCOCO train 2014 (400) } & 90.75  & 9.25  & 91.25 &  8.75 & 73.5 & 8.25 & 2.75  & 6.75  & 0.5 & 0.5    & 7.75 \\
    \textbf{MSCOCO val 2014 (90)} & 97.77 & 2.3 & 94.44 & 8.75  & 84.44  & 8.88  & 3.33  & 1.11  & 0  & 0   & 2.22 \\
  \textbf{Flickr30K Entities 2015 (510)} & 78.24 & 21.76  & 73.73 & 26.27 & 64.00 & 4.3 & 1.7  & 1.7  & 0.7   & 1.18  & 26.45 \\
  \hline
    \end{tabular}%
    }
    \vspace{-5mm}
\end{table*}

In the MS COCO dataset, one challenging aspect is that the S or O can be singular, plural, or referring to the scene. This means that one S could map to multiple boxes in the image. For example, ``people'' maps to multiple boxes of ``person''.  Furthermore, this case could exist for both the S and the O.  In cases where either S or O is plural, the bounding box assigned is the union of all candidate bounding boxes in  $\mathbf{b}_S^i$.   The grounding then proceeds as follows.

\textbf{$<$S,P$>$ facts: }

(1) If the computed $\mathbf{b}_S^i = \varnothing$ for the given $\mathbf{f}_l^i$, then $\mathbf{f}_l^i$  fails to ground and is discarded.  

(2) If S singular,  $\mathbf{f}_v^i$ is the image region that with the largest candidate bounding box in  $\mathbf{b}_S^i$.

(3) If S is plural, $\mathbf{f}_v^i$ is the image region that with union of the candidate bounding boxes in  $\mathbf{b}_S^i$.

\textbf{$<$S,P, O$>$ facts: }

(1) If $\mathbf{b}_S^i =  \varnothing$ and  $\mathbf{b}_O^i = \varnothing$,  $\mathbf{f}_l^i$ fails to ground and is ignored.

(2) If $\mathbf{b}_S^i\neq \varnothing$ and  $\mathbf{b}_O^i \neq \varnothing$, then bounding boxes are assigned to S and O such that the distance between them is minimized (though if S or O is plural, the assigned bounding box is the union of all bounding boxes for $\mathbf{b}_S^i$ or $\mathbf{b}_O^i$ respectively), and the grounding is assigned the union of the bounding boxes assigned to S and O.

(3) If either $\mathbf{b}_S^i = \varnothing$ or  $\mathbf{b}_O^i = \varnothing$, then a bounding box is assigned to the present object (the largest bounding box if singular, or the union of all bounding boxes if plural).  If the area of this region compared to the area of the whole scene is greater than a threshold $th=0.3$, then the $\mathbf{f}_v^i$ is   associated to the whole image of the scene.  Otherwise,  $\mathbf{f}_l^i$ fails to ground and is ignored.

\textbf{Grounding: Flickr30K dataset}
The main difference in Flickr30K is that for each entity phrase in a sentence, there is a box in the image. This means there is no need to have cases for single and plural. Since in this case, the word ``men'' in the sentence will be associated with the set of boxes referred to by  ``men'' in the sentences.  We union these boxes for plural words as one candidate box for ``men''

We can also use the information that the object box has to refer to a word that is after the subject word, since subject usually occurs earlier in the sentence compared to object.  We union these boxes for plural words.

\textbf{$<$S,P$>$ facts: }

If  the computed $\mathbf{b}_S^i = \varnothing$ for the given $\mathbf{f}_l^i$, then $\mathbf{f}_l^i$  fails to ground and is discarded. Otherwise, the fact is assigned to the largest candidate box in if there are multiple boxes.

\textbf{$<$S,P, O$>$ facts: }
$<$S,P, O$>$ facts are handled very similar to MSCOCO dataset with two main differences.

a) The candidate boxes are computed as described for the case of Flickr30K dataset. 

b) All cases are handled as single case, since even plural words are assigned one box based on the nature of the annotations in this dataset.


\vspace{-2mm}
\section{Experiments}
\vspace{-2mm}
\label{sec_Expr}

\subsection{Human Subject Evaluation}


We propose  three questions to evaluate each annotation: (Q1) Is the extracted fact correct (Yes/No)?  The purpose of this question is to evaluate errors captured by the first step, which extracts facts by Sedona or Clausie. (Q2) Is the fact located in the image (Yes/No)?  In some cases, there might be a fact  mentioned in the caption that does not exist in the image and is mistakenly considered as an annotation. (Q3) How accurate is the box assigned to a given fact (a to g)?  a (about right), b (a bit big), c (a bit small), d (too small), e (too big), f (totally wrong box), g (fact does not exist or other). 
Our instructions on these questions to the participants can be found in this anonymous url~\cite{safaevalinstructions15}.


We evaluate these three questions for the facts that were successfully assigned a box in the image, because the main purpose of this evaluation is to measure the usability of the collected annotations as training data for our model. We created an Amazon Mechanical Turk form to ask these three questions.  So far, we collected a total of 10,786 evaluation responses, which are an evaluation of 3,595  ($\mathbf{f}_v, \mathbf{f}_l$)  pairs (3 responses/ pair). 
Table~\ref{tbl:afaf_eval_mturk} shows the evaluation results, which indicate that the  data is useful for training, since$\approx$83.1\% of them are correct facts with boxes that are either about right, or  a bit big or small (a,b,c). We further some evaluation responses that we collected from volunteer researchers in Table~\ref{tbl:afaf_eval_mturk} showing similar results. 

\begin{figure}[b!]
\vspace{-1mm}
  \centering
 \hspace{-6mm} \includegraphics[width=0.52\textwidth]{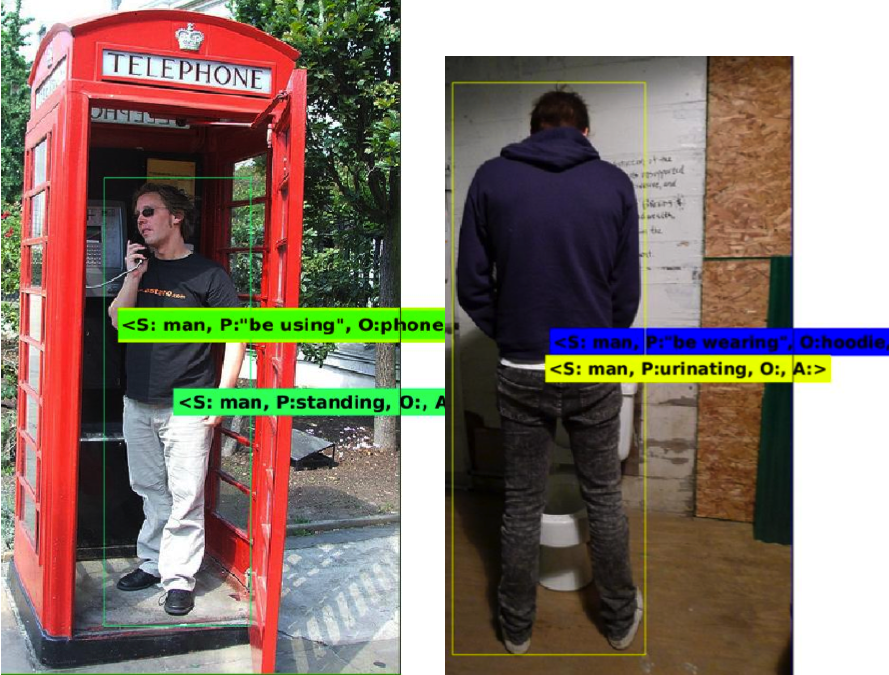} 
   \vspace{-4mm}
\caption{Se6eral Facts successfully extracted by our method from two MS COCO scenes}
  \label{safa_pos_Ex}
  \includegraphics[width=0.45\textwidth]{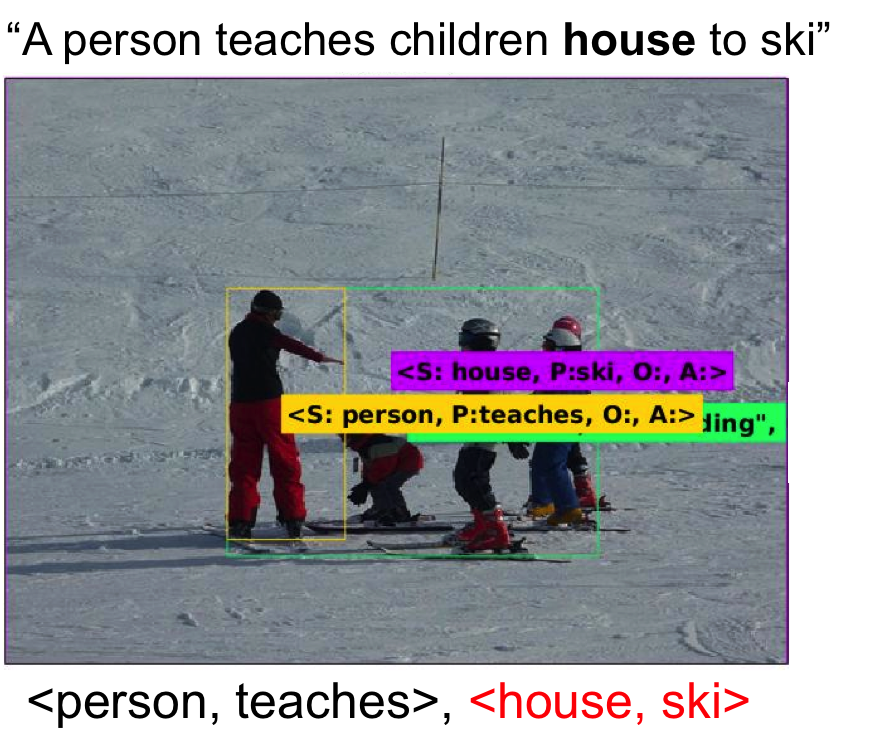}
   \vspace{-2mm}
 \caption{An example where one of the extracted facts are not correct due to a spelling mistake}
 \label{safa_neg_Ex}
  \vspace{-5mm}
\end{figure}

Fig.~\ref{safa_pos_Ex} shows some  successful qualitative results that include four extracted structured facts from MS COCO dataset (e.g., $<$person, using, phone$>$, $<$person, standing$>$, etc).  Fig~\ref{safa_neg_Ex} also show a negative example where there is a wrong fact among the extracted facts (i.e., $<$house, ski$>$). The main reason for this failure case is that ``how'' is mistyped as ``house''; see Fig~\ref{safa_neg_Ex}. The supplementary materials includes all the captions of these examples and also  additional qualitative examples.

\subsection{Hardness Evaluation of the collected data}

In order to study how the method behave in both easy and hard examples. This section present statistics of the successfully extracted facts and relate it to the hardness of the extraction of these facts.  We start by defining hardness of an extracted fact in our case and its dependency on the fact type.  Our method collect both second- and third-order facts.  We refer to candidate subjects as all instances of the entity in the image that match the subject type  of either a second-order fact $<$S,P$>$ or a third-order fact $<$S,P,O$>$. We refer to candidate objects as all instances  in the image that match the object type  of a third-order fact $<$S,P,O$>$. The selection of the candidate subjects and candidate objects is a part of our method that we detailed in Sec~\ref{sec_locate_facts}. We define the hardness for second order facts by the number of candidate subjects and the hardness of third order facts by the number of candidate subjects multiplied by the number of candidate objects.  


In Fig~\ref{fgmturk_all} and~\ref{fgmturkq31},  the Y axis is the number of facts for each bin. The X axis shows the bins that correspond to hardness that we defined for both second and third order fats. Figure~\ref{fgmturk_all} shows a histogram of the difficulties for all Mturk evaluated examples including both the successful and the failure cases. Figure~\ref{fgmturkq31} shows a similar histogram but for but for subset of facts  verified by the Turkers with Q3 as (about right). The figures show that the method is able to handle difficulty cases even with more than 150 possibilities for grounding. We show these results broken out for MSCOCO and Flickr30K Entities datasets and  for each fact types in the supplementary materials.

\begin{figure}[ht!]
\vspace{-4mm}
  \centering
      \includegraphics[width=0.5\textwidth,height=0.3\textwidth]{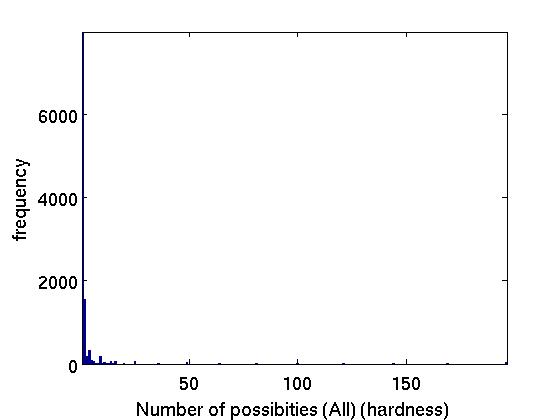}
        \caption{(All MTurk Data) Hardness histogram after candidate box selection using our method}
        \label{fgmturk_all}
      \includegraphics[width=0.5\textwidth,height=0.3\textwidth]{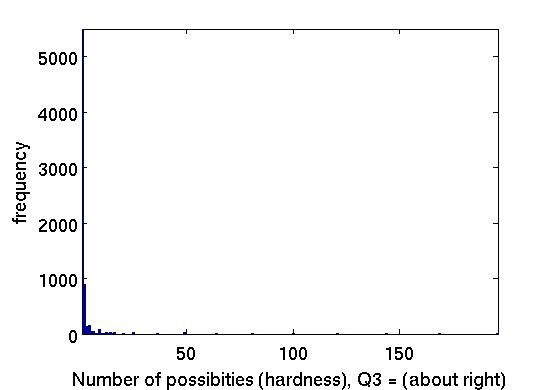}
  \caption{(MTurk Data with Q3=about right)Hardness histogram  after our candidate box selection}
  \label{fgmturkq31}
\end{figure}



\vspace{-2mm}
\section{Conclusion}
\vspace{-2mm}
We present a new method whose main purpose to collect visual fact annotation by a language approach. The collected data help train visual system systems on the fact level with the diversity of facts captured by any fact described by an image caption. We showed the effectiveness of the proposed methodology by extracting hundreds of thousands of fact-level annotations from MSCOCO and Flickr30K datasets.  We verified and analyzed the collected data and showed that more than 80\% of the collected data are good for training visual systems.



\bibliography{egbib}
\bibliographystyle{acl2016}
\clearpage
\section*{Supplementary Materials}

This document includes several qualitative examples of high order facts automatically collected from MSCOCO dataset and Flickr30 Dataset as well. 

1) Qualitative Examples

2) Statistic on the COllected data

3) Extracted Facts and Statistic on them using Sedona for each of MSCOCO train, MSCOCO validation, and Flickr30K datasets. There is an attached zip file for each dataset.  





\section{SAFA Qualitative Results}
\subsection*{SAFA successful cases}

On the left is the input, and on the right is the output.

\begin{figure*}[h!]
  \centering
    {%
      \includegraphics[width=1.0\textwidth]{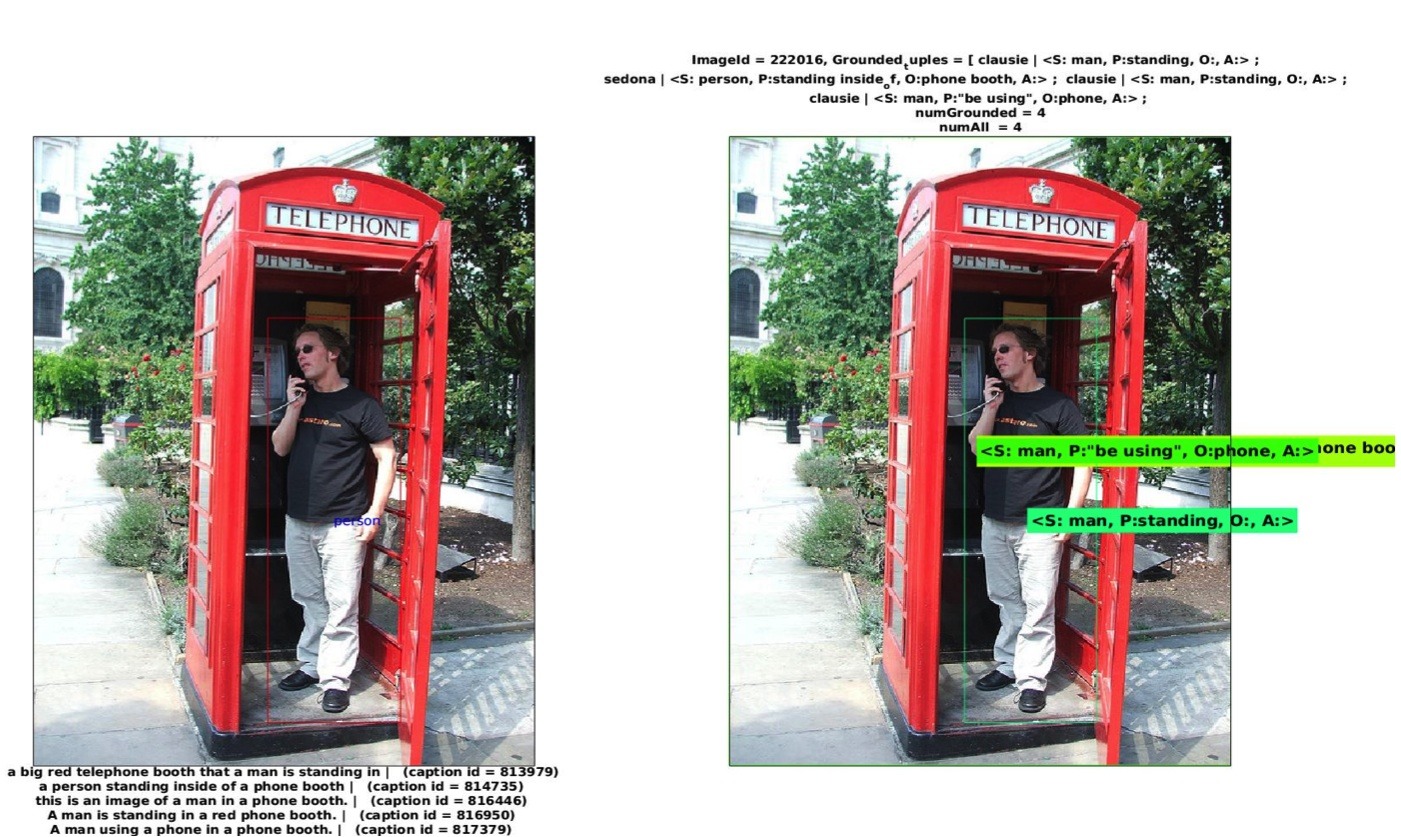}}
  \caption{Example 1 from MS COCO dataset}
\end{figure*}

\begin{figure*}[h!]
  \centering
    {%
      \includegraphics[width=1.0\textwidth]{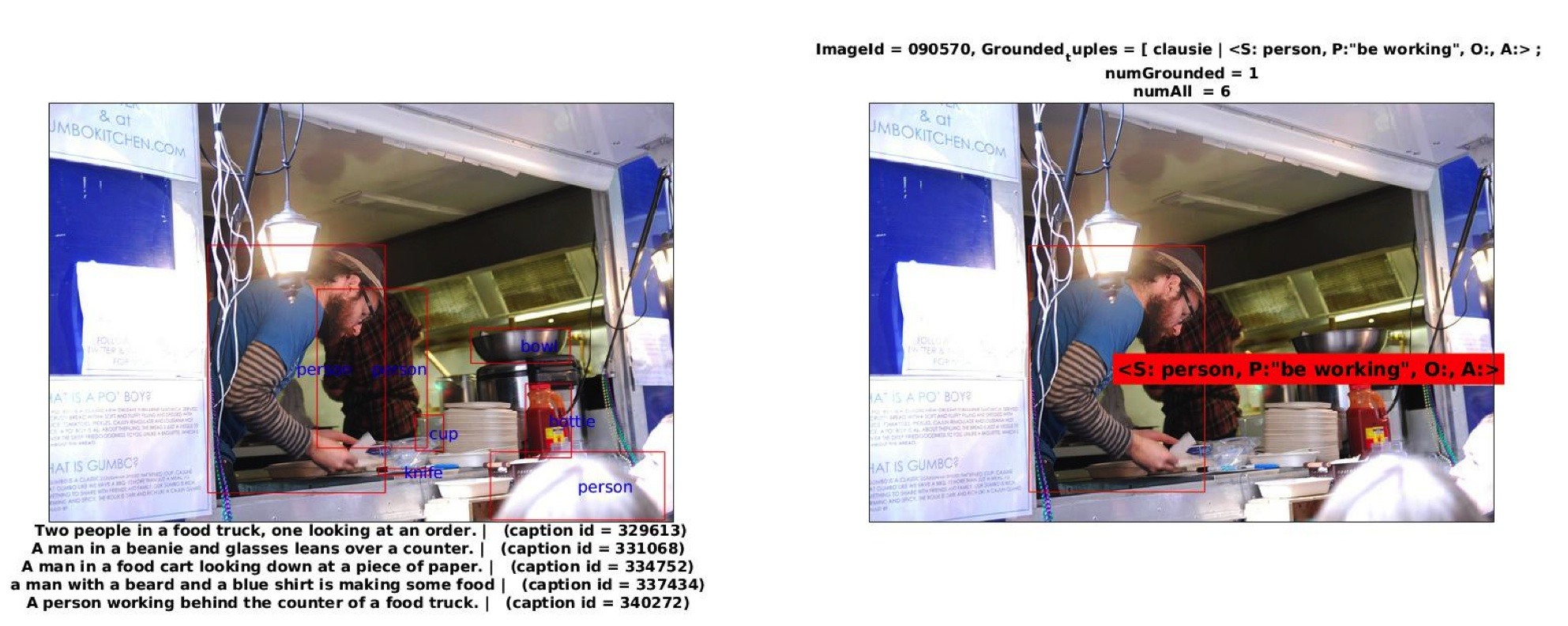}}
  \caption{Example 2 from MS COCO dataset}
\end{figure*}

\begin{figure*}[h!]
  \centering
    {%
      \includegraphics[width=1.0\textwidth]{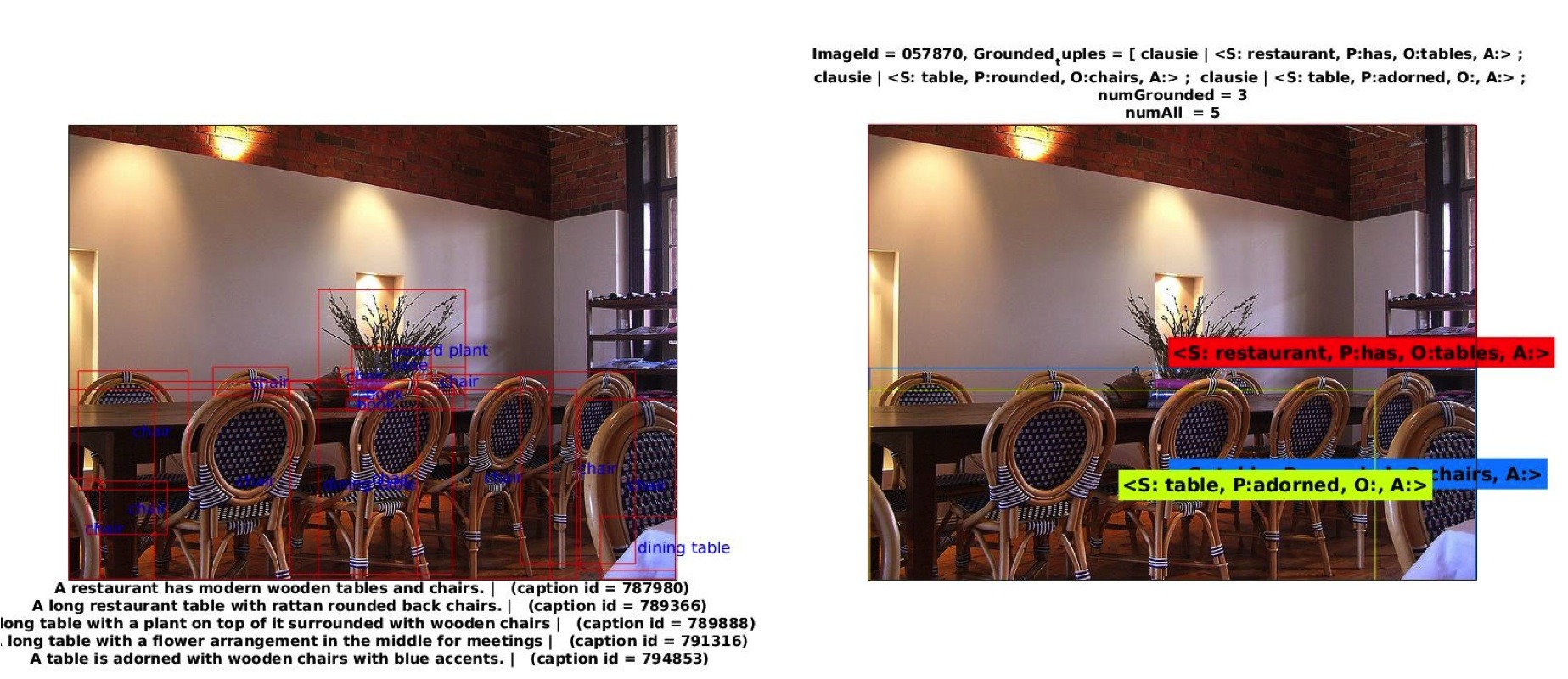}}
  \caption{Example 3 from MS COCO dataset}
\end{figure*}

\begin{figure*}[h!]
  \centering
    {%
      \includegraphics[width=1.0\textwidth]{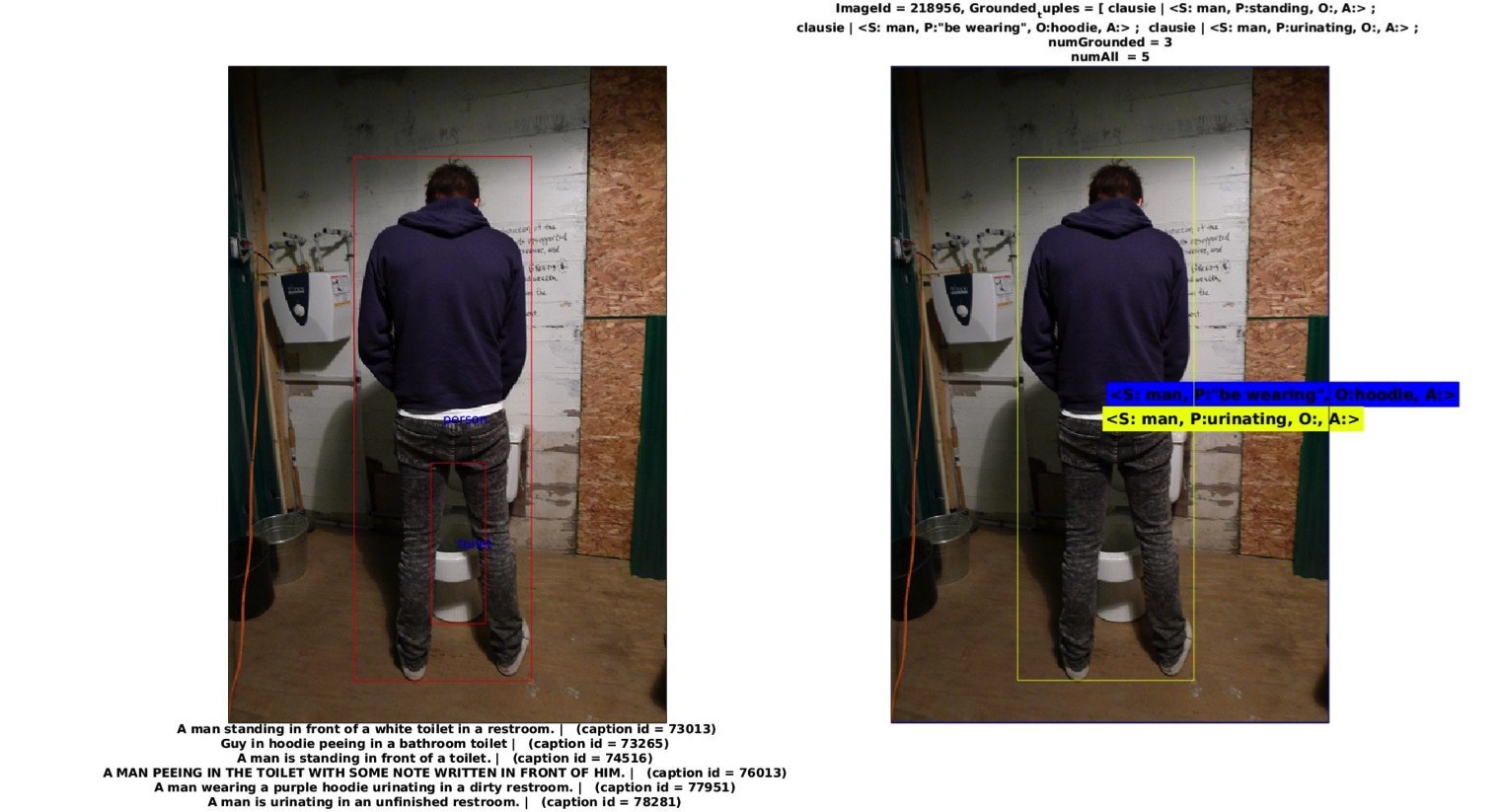}}
  \caption{Example 4  from MS COCO dataset}
\end{figure*}

\clearpage

\subsection*{SAFA failure cases}

\begin{figure*}[h!]
  \centering
    {%
      \includegraphics[width=1.0\textwidth]{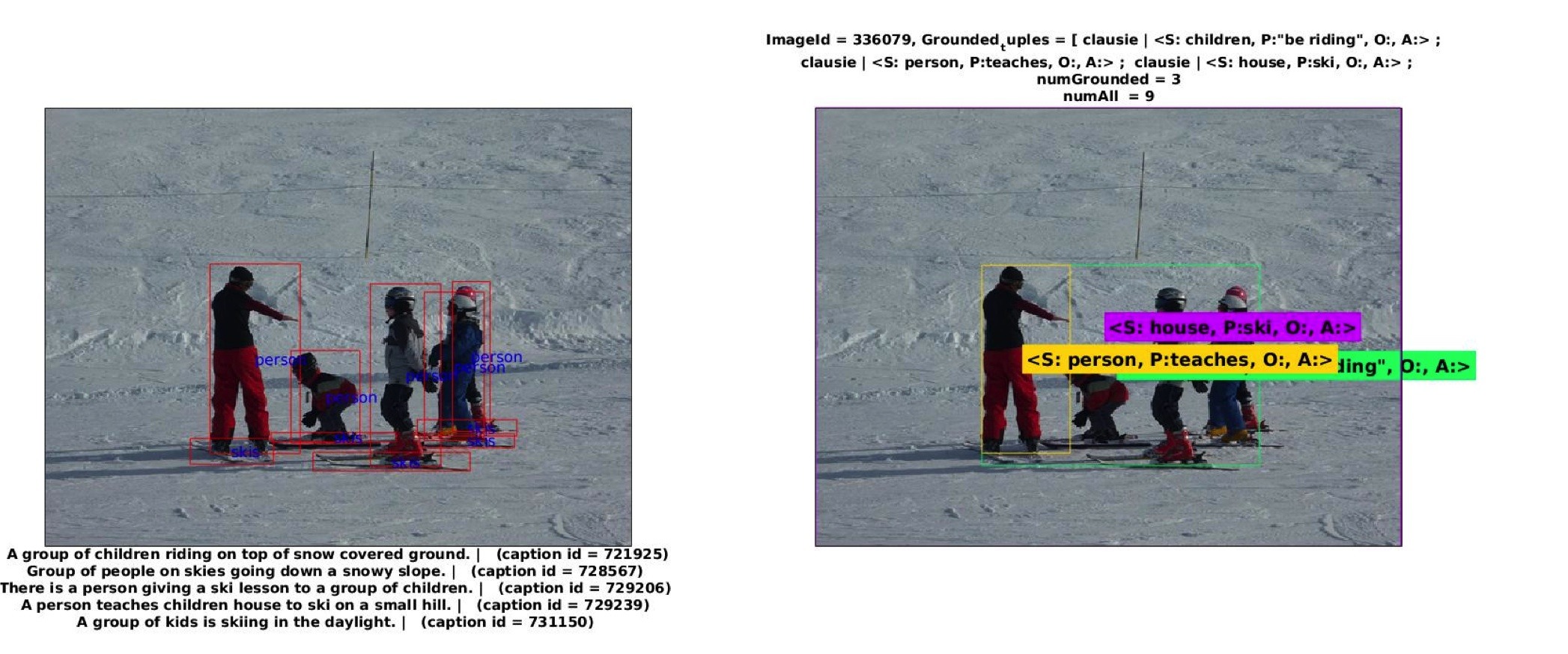}}
  \caption{Failure Case 1 from MS COCO dataset:  ($<S:$ house$,$ $P:$ ski $>$), due to a spelling mistake in the statement ``A person teaches children house to ski''}
\end{figure*}

\begin{figure*}[h!]
  \centering
    {%
      \includegraphics[width=1.0\textwidth]{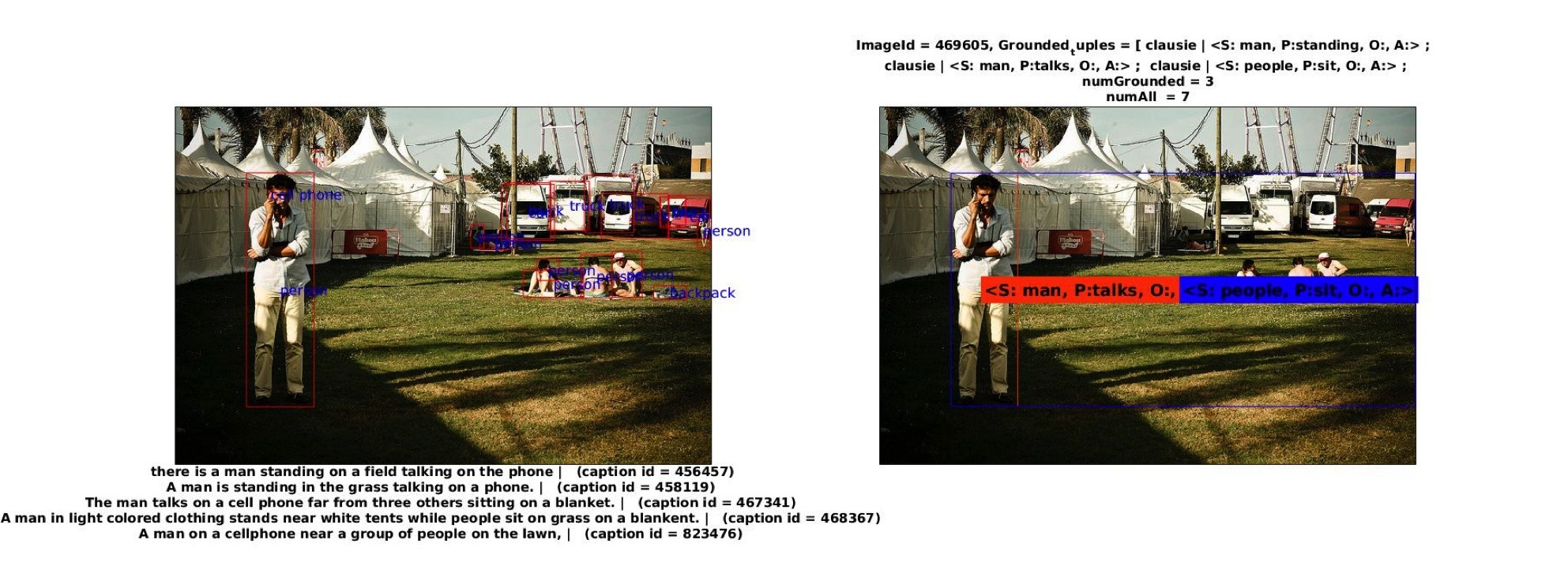}}
  \caption{Failure Case 2 from MS COCO dataset: the bounding box of $<S:$ people$, P:$ sit $>$ covers the man that is not sitting. This happens since our heuristics in this case unions all annotations that could be a person since people is a group of persons}
\end{figure*}

\subsection{Statistics on collected data (MTurk  subset)}

In order to study how the method behave in both easy and hard examples. This section present statistics of the successfully extracted facts and relate it to the hardness of the extraction of these facts.  We start by defining hardness of an extracted fact in our case and its dependency on the fact type.  Our method collect both second- and third-order facts.  We refer to candidate subjects as all instances of the entity in the image that match the subject type  of either a second-order fact $<$S,P$>$ or a third-order fact $<$S,P,O$>$. We refer to candidate objects as all instances  in the image that match the object type  of a third-order fact $<$S,P,O$>$. The selection of the candidate subjects and candidate objects is a part of our method that we detailed in Sec.~4 in the paper. We define the hardness for second order facts by the number of candidate subjects and the hardness of third order facts by the number of candidate subjects multiplied by the number of candidate objects.  


In all the following figures the Y axis is the number of facts for each bin. The X axes are bins that correspond to 

(1) the number of candidate subjects  for second and third order facts.

(2) the number of candidate objects  for  third order facts. 

(3) the number of candidate objects multiplied by number of candidate subjects for  third-order facts (which is all possible pairs of entities in an image that match the given $<$S,P,O$>$ fact)

We here show the number of possible candidates as the level of difficulty/hardness on the x axis, y axis is the number of facts in each case. 

\subsubsection{Statistics on both all MTurk data compared to the subset where Q3 =  about right for each dataset}
The following figures show statistics on the facts verified by MTurkers (from the union of all datasets). Figure~\ref{fgmturk_all} shows a histogram of the difficulties for all Mturk evaluated examples.  Figure~\ref{fgmturkq31} shows a similar histogram but for but for subset of facts  verified by the Turkers by Q3 as (about right) by MTurker. The figures show that the method is able to handle difficulty cases even with more than 150 possibilities for grounding.

\begin{figure}[h!]
  \centering
      \includegraphics[width=0.5\textwidth]{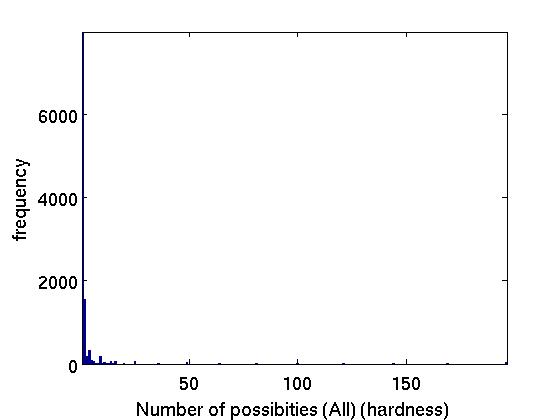}
        \caption{(All MTurk Data) Number or Possibilities for grounding each item after Natural Language Processing}
        \label{fgmturk_all}
\end{figure}

\begin{figure}[h!]
  \centering
      \includegraphics[width=0.5\textwidth]{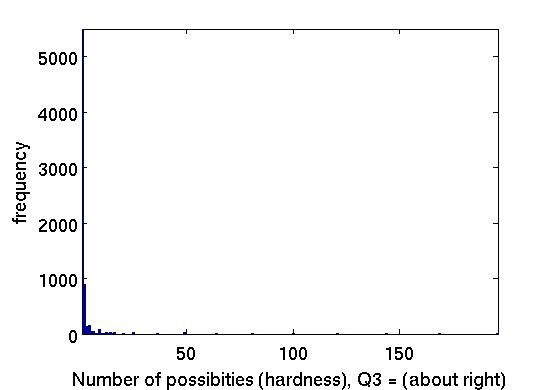}
  \caption{(MTurk Data with Q3 = about right) Number or Possibilities for grounding each item after Natural Language Processing}
  \label{fgmturkq31}
\end{figure}
\subsubsection{Statistics on Q3 =  about right for each dataset}
\begin{figure}[h!]
  \centering
  \includegraphics[width=0.5\textwidth]{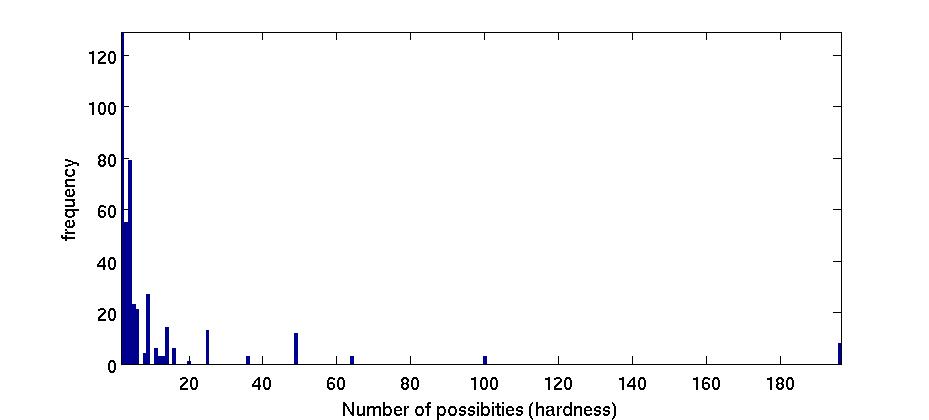}
      \includegraphics[width=0.5\textwidth]{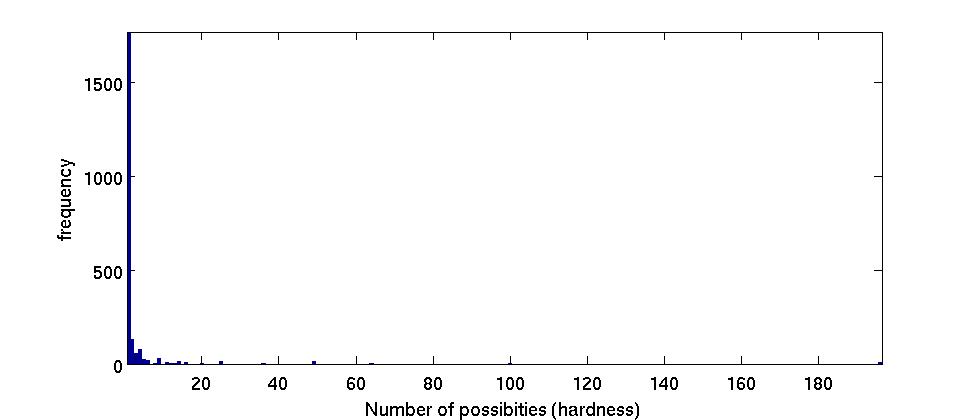}
  \caption{(Flickr30K Dataset)  Hardness of Automatically collected perfect examples (Q3 = about right). (Bottom is the same figure starting from hardness starting from 2 candidates and more)}
\end{figure}

\begin{figure}[h!]
  \centering
  \includegraphics[width=0.5\textwidth]{hardcases_Q3_1_Flick30K_.jpg}
      \includegraphics[width=0.5\textwidth]{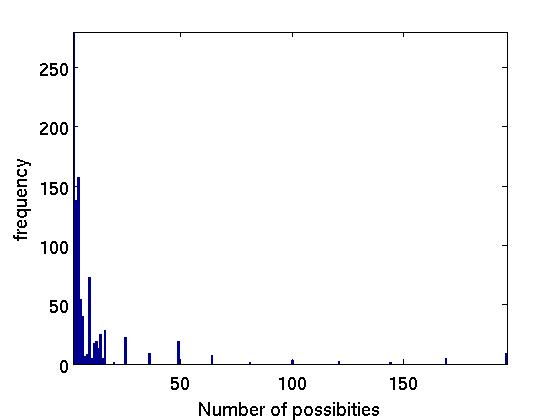}
 \caption{(MSCOCO Dataset)  Hardness of Automatically collected perfect examples (Q3 = about right)(Bottom is the same figure starting from hardness from 2 candidates and more)}
\end{figure}

\section{SAFA collected data statistics (the whole 380K collected data)}
This section shows the number of candidate subject and object statistics for all successfully grounded facts for all MS COCO  (union of training and validation subsets) and Flickr30K datasets.   SAFA collects second- and third-order facts.  We refer to candidate subjects as all instances of the entity that match the subject  of either a second-order fact $<$S,P$>$ or a third-order fact $<$S,P,O$>$. We refer to candidate objects as all instances of the entity that match the object  of a third-order fact $<$S,P,O$>$. The selection of the candidate subjects and candidate objects is a part of our method that we detailed in this paper. Our method was designed to achieve high precision such that the grounded facts are as accurate as possible as we showed in our experiments. 

In all the following figures the Y axis is the number of facts for each bin. The X axes are bins that correspond to 

(1) the number of candidate subjects  for second and third order facts.

(2) the number of candidate objects  for  third order facts. 

(3) the number of candidate objects multiplied by number of candidate subjects for  third-order facts (which is all possible pairs of entities in an image that match the given $<$S,P,O$>$ fact)

\subsection{Second-Order Facts $<$S,P$>$: Candidate Subjects}
\begin{figure}[h!]
  \centering
      \includegraphics[width=0.5\textwidth]{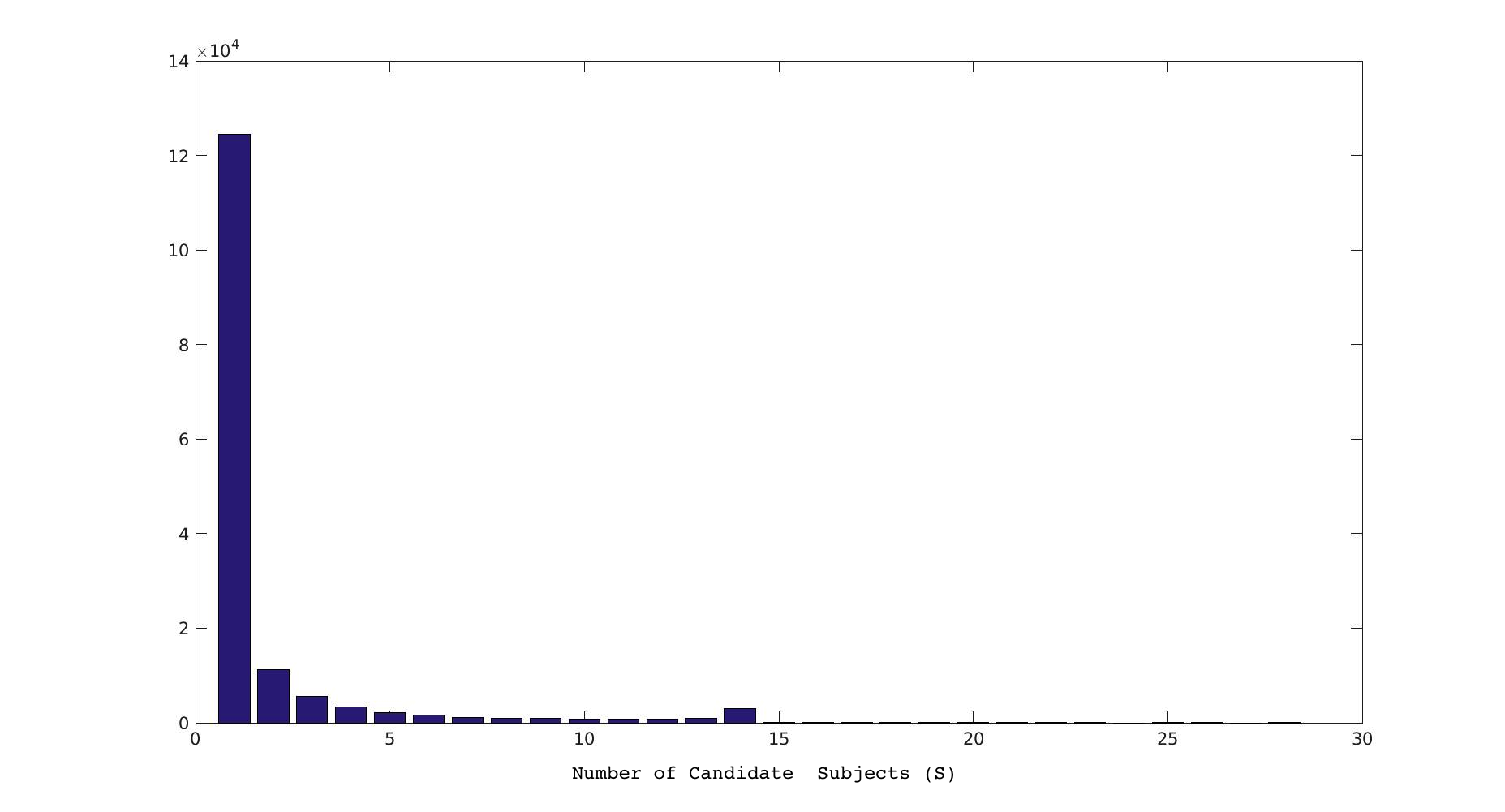}
  \caption{(COCO Dataset) Automatically Collected Second-Order Facts ($<$S,P$>$) (Y axis is number of facts)}
\end{figure}

\begin{figure}[h!]
  \centering
      \includegraphics[width=0.5\textwidth]{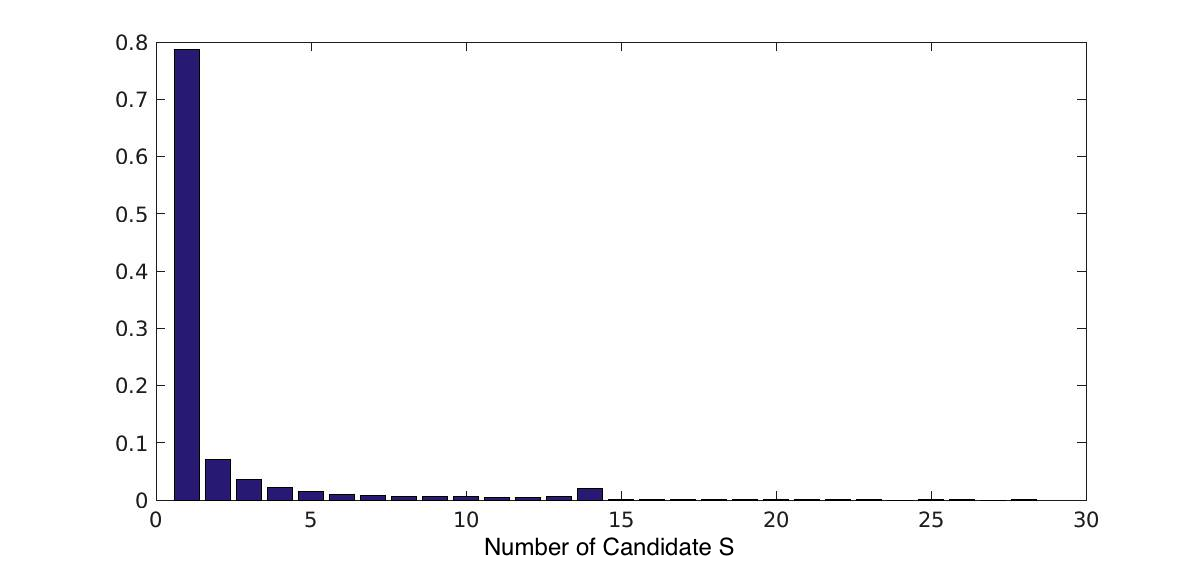}
  \caption{(COCO Dataset) Automatically Collected Second-Order Facts ($<$S,P$>$) (Y axis is ratio of facts)}
\end{figure}

\begin{figure}[h!]
  \centering
      \includegraphics[width=0.5\textwidth]{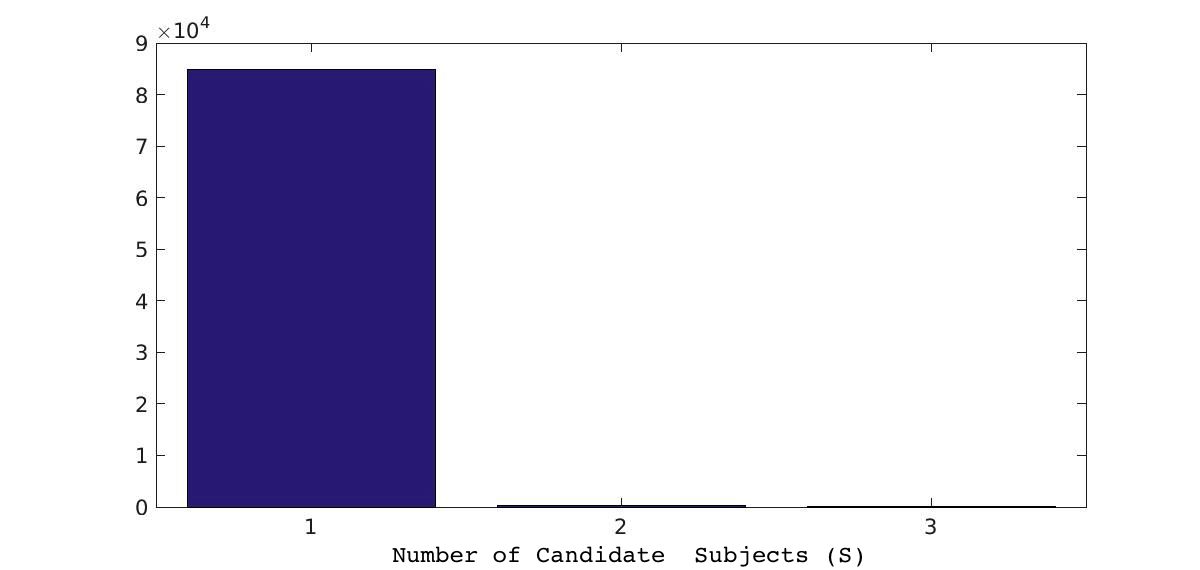}
  \caption{(Flickr30K Dataset) Automatically Collected Second-Order Facts ($<$S,P$>$) (Y axis is the number  of facts) }
\end{figure}

\begin{figure}[h!]
  \centering
      \includegraphics[width=0.5\textwidth]{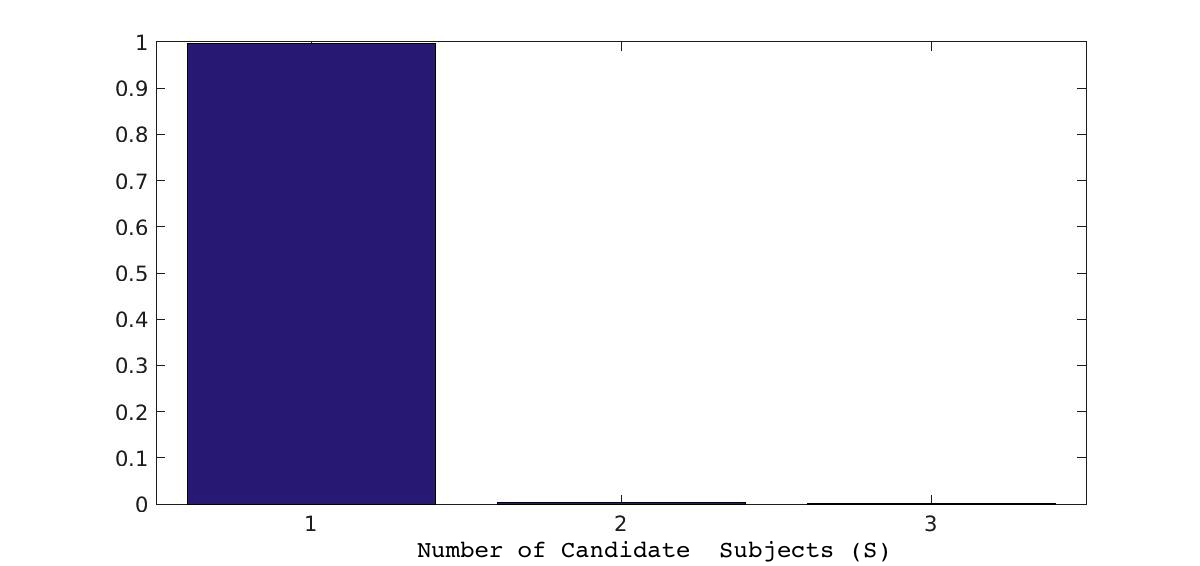}
  \caption{(Flickr30K Dataset) Automatically Collected Second-Order Facts ($<$S,P$>$) (Y axis is ratio of facts)}
\end{figure}

\clearpage
\subsection{Third-Order Facts $<$S,P,O$>$:  Candidate Subjects}

\begin{figure}[h!]
  \centering
      \includegraphics[width=0.5\textwidth]{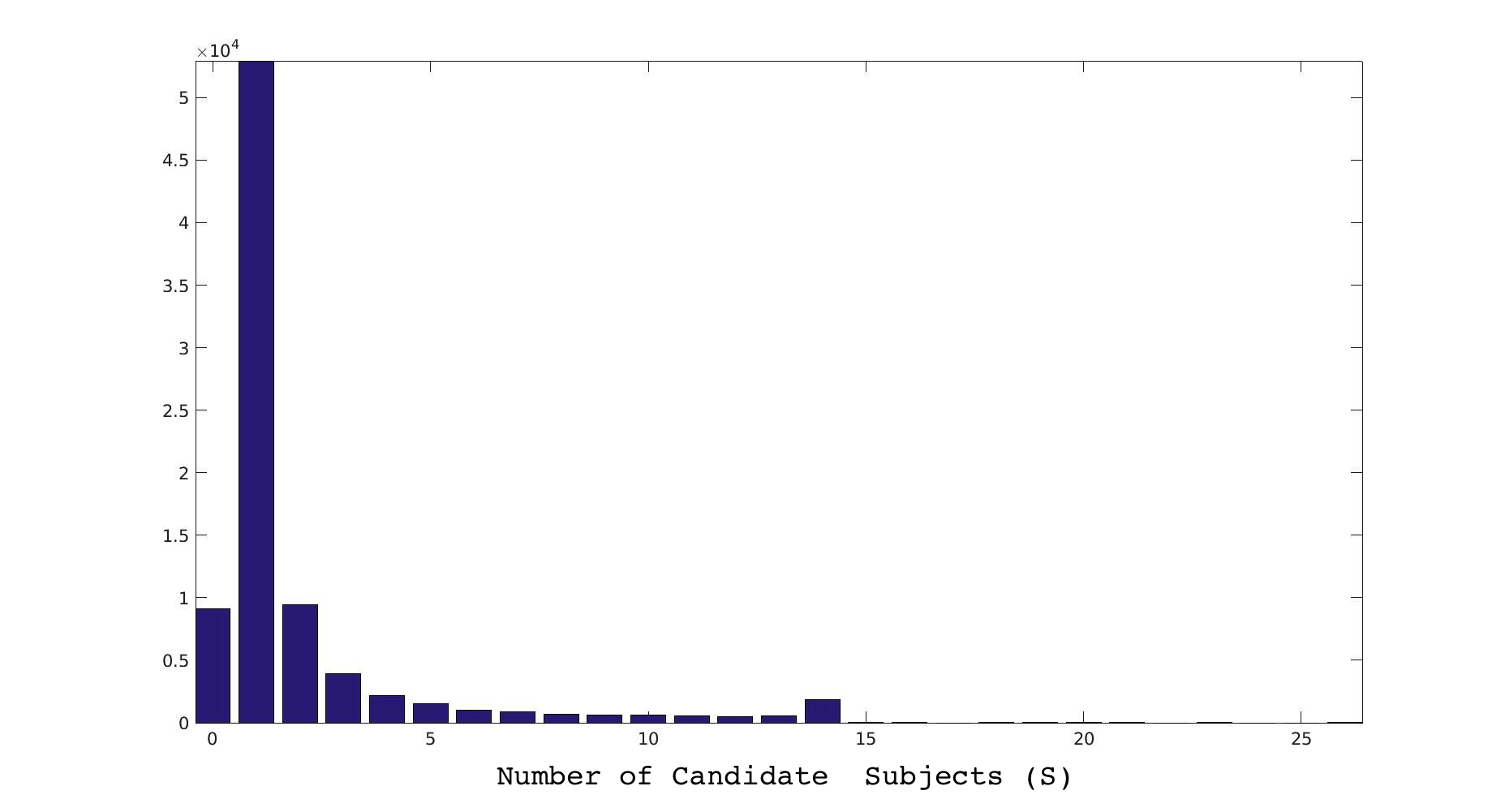}
  \caption{(COCO Dataset) Automatically Collected Third-Order Facts ($<$S,P,O$>$) (Y axis is the number  of facts) }
\end{figure}

\begin{figure}[h!]
  \centering
      \includegraphics[width=0.5\textwidth]{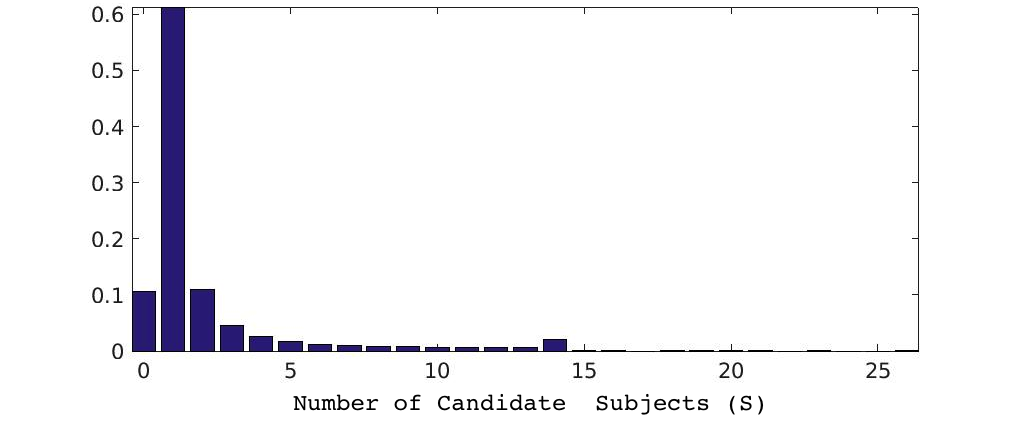}
  \caption{(COCO Dataset) Automatically Collected Third-Order Facts ($<$S,P,O$>$) (Y axis is the ratio  of facts) }
\end{figure}

\begin{figure}[h!]
  \centering
      \includegraphics[width=0.5\textwidth]{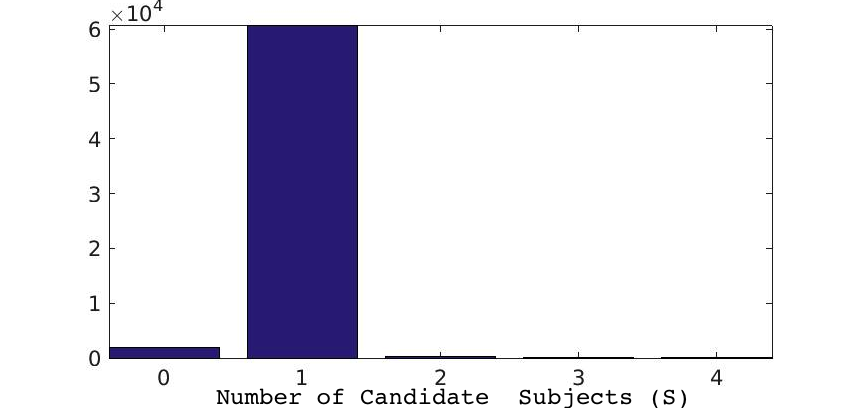}
  \caption{(Flickr30K Dataset) Automatically Collected Third-Order Facts ($<$S,P,O$>$) (Y axis is the number  of facts) }
\end{figure}

\begin{figure}[h!]
  \centering
      \includegraphics[width=0.5\textwidth]{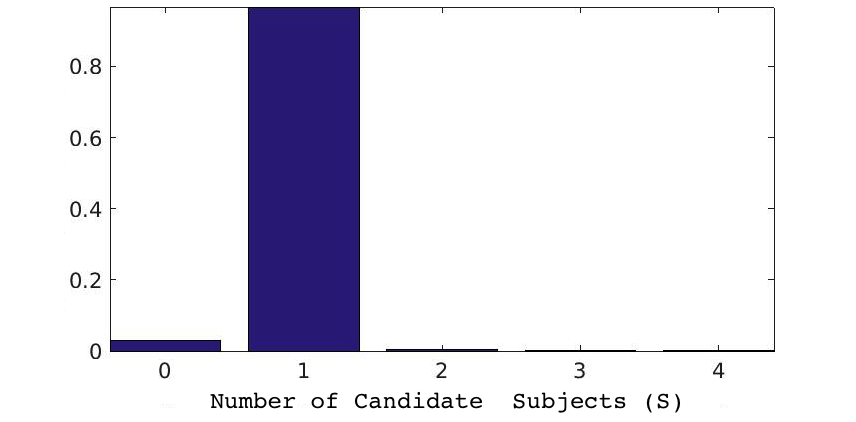}
  \caption{(Flickr30K Dataset) Automatically Third-Order Facts ($<$S,P,O$>$)  (Y axis is the ratio  of facts) }
\end{figure}

\clearpage

\subsection{Third-Order Facts $<$S,P,O$>$:  Candidate Objects}

\begin{figure}[h!]
  \centering
      \includegraphics[width=0.5\textwidth]{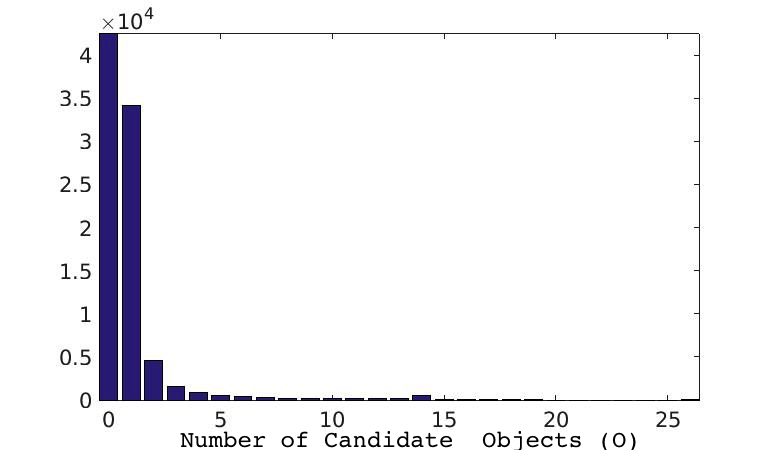}
  \caption{(COCO Dataset) Automatically Collected Third-Order Facts ($<$S,P,O$>$) (Y axis is the number  of facts) }
\end{figure}

\begin{figure}[h!]
  \centering
      \includegraphics[width=0.5\textwidth]{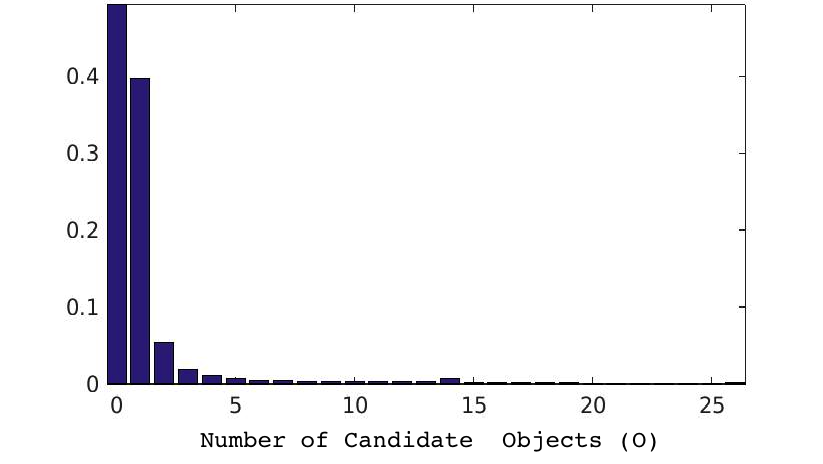}
  \caption{(COCO Dataset) Automatically  Collected Third-Order Facts ($<$S,P,O$>$) (Y axis is the ratio  of facts) }
\end{figure}

\begin{figure}[h!]
  \centering
      \includegraphics[width=0.5\textwidth]{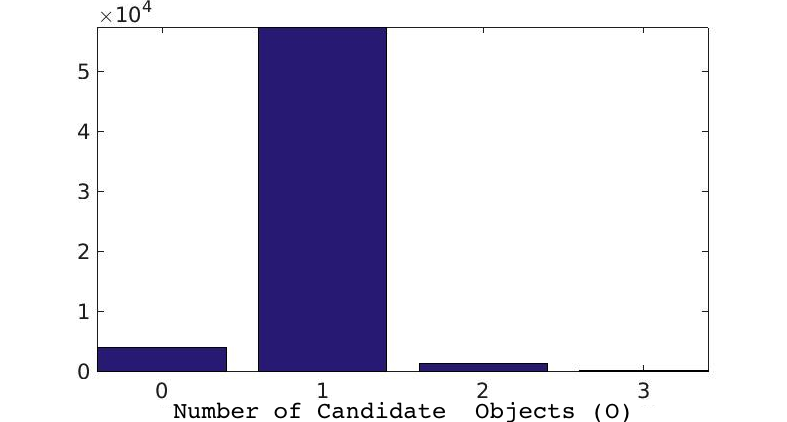}
  \caption{(Flickr30K Dataset) Automatically  Collected Third-Order Facts ($<$S,P,O$>$) (Y axis is the number  of facts) }
\end{figure}

\begin{figure}[h!]
  \centering
      \includegraphics[width=0.5\textwidth]{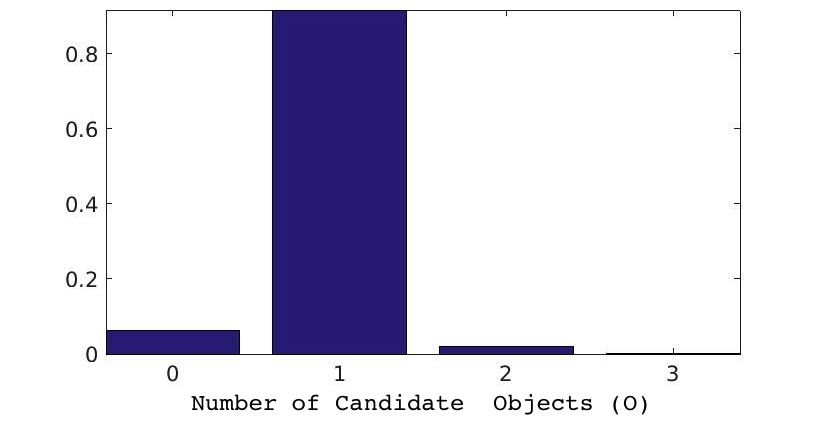}
  \caption{(Flickr30K Dataset) Automatically  Collected Third-Order Facts ($<$S,P,O$>$)  (Y axis is the ratio  of facts) }
\end{figure}

\clearpage

\subsection{Third-Order Facts $<$S,P,O$>$:  Candidate Subjects/Object Combination}

\begin{figure}[h!]
  \centering
      \includegraphics[width=0.5\textwidth]{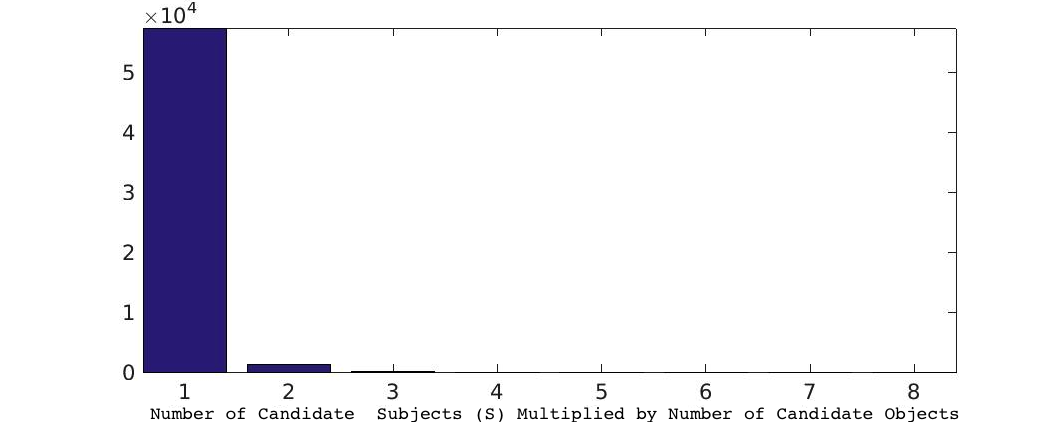}
  \caption{(COCO Dataset) Automatically Second-Order Facts ($<$S,P,O$>$) (Y axis is the number  of facts) }
\end{figure}

\begin{figure}[h!]
  \centering
      \includegraphics[width=0.5\textwidth]{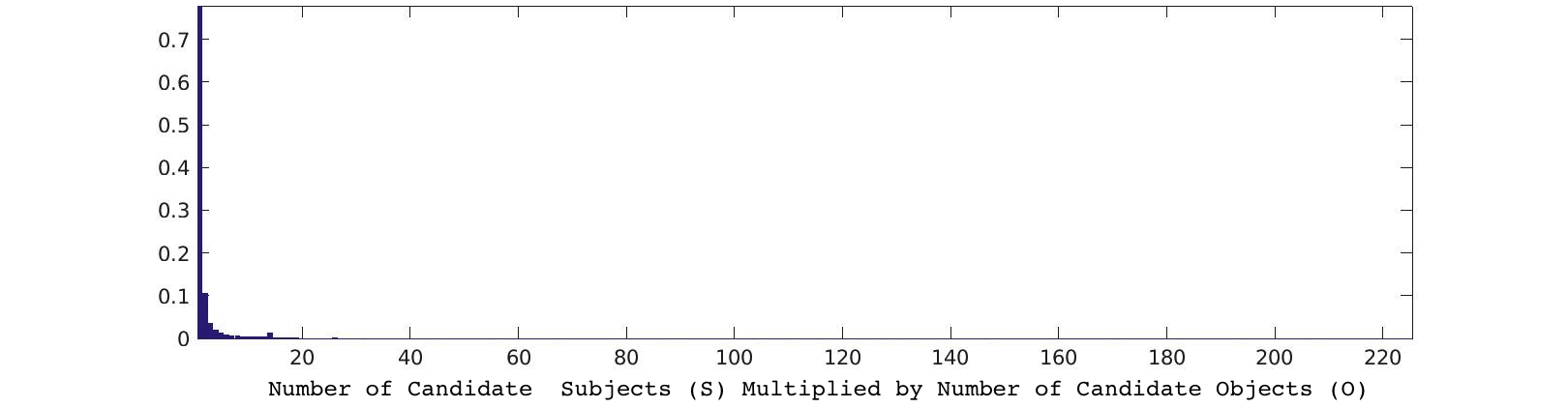}
  \caption{(COCO Dataset) Automatically Second-Order Facts ($<$S,P,O$>$) (Y axis is the percentage  of facts) }
\end{figure}

\begin{figure}[h!]
  \centering
      \includegraphics[width=0.5\textwidth]{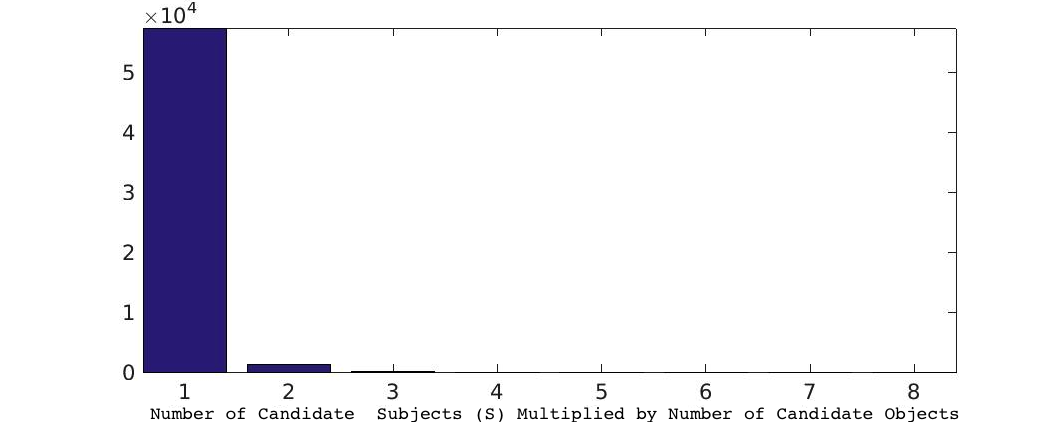}
  \caption{(Flickr30K Dataset) Automatically Second-Order Facts ($<$S,P,O$>$) (Y axis is the number  of facts) }
\end{figure}

\begin{figure}[h!]
  \centering
      \includegraphics[width=0.5\textwidth]{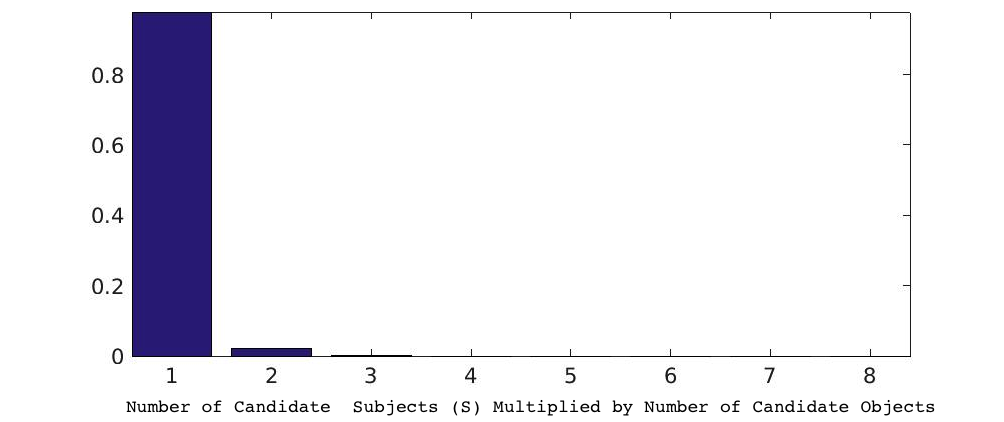}
  \caption{(Flickr30K Dataset) Automatically Second-Order Facts ($<$S,P$>$) (Y axis is the ratio  of facts) }
\end{figure}
\clearpage

\end{document}